\documentclass[pdflatex,sn-mathphys-ay]{sn-jnl}

\usepackage{graphicx}%
\usepackage{multirow}%
\usepackage{amsmath,amssymb,amsfonts}%
\usepackage{amsthm}%
\usepackage{mathrsfs}%
\usepackage[title]{appendix}%
\usepackage{xcolor}%
\usepackage{textcomp}%
\usepackage{manyfoot}%
\usepackage{booktabs}%
\usepackage{algorithm}%
\usepackage{algorithmicx}%
\usepackage{algpseudocode}%
\usepackage{listings}%

\theoremstyle{thmstyleone}%
\theoremstyle{thmstyletwo}%
\newtheorem{remark}{Remark}%

\theoremstyle{thmstylethree}%

\usepackage{bookmark}
\usepackage{bbm}
\newcommand{\R}{\mathbb{R}}

\colorlet{Review}{black}
\newcommand{\review}[1]{#1}

\begin{document}

\title{Neural non-canonical Hamiltonian dynamics for long-time simulations}


\author[1,2]{\fnm{Clémentine} \sur{Courtès}}
\author[1,2]{\fnm{Emmanuel} \sur{Franck}}
\author[3]{\fnm{Michael} \sur{Kraus}}
\author[1,2]{\fnm{Laurent} \sur{Navoret}}
\author*[4]{\fnm{Léopold} \sur{Trémant}}\email{leopold.tremant@univ-artois.fr}

\affil[1]{\orgdiv{Institut de Recherche Mathématique Avancée, UMR 7501} \orgname{Université de Strasbourg et CNRS}, \orgaddress{\street{7~rue René Descartes} \postcode{672000} \city{Strasbourg}, \country{France}}}

\affil[2]{\orgname{INRIA Nancy-Grand Est}, \orgdiv{MACARON Project}, \orgaddress{\city{Strasbourg}, \country{France}}}

\affil[3]{\orgname{Max-Planck-Institut für Plasmaphysik}, \orgaddress{\city{Garching bei München}, \country{Germany}}}

\affil[4]{\orgname{Univ. Artois}, \orgdiv{UR 2426, Laboratoire de Mathématiques de Lens (LML)}, \orgaddress{\postcode{F-62300}, \city{Lens}, \country{France}}}


\abstract{%
This work focuses on learning non-canonical Hamiltonian dynamics from data, where long-term predictions require the preservation of structure both in the learned model and in numerical schemes. Previous research focused on either facet, respectively with a potential-based architecture and with degenerate variational integrators, but new issues arise when combining both. In experiments, the learnt model is sometimes numerically unstable due to the gauge dependency of the scheme, rendering long-time simulations impossible. In this paper, we identify this problem and propose two different training strategies to address it, either by directly learning the vector field or by learning a time-discrete dynamics through the scheme. Several numerical test cases assess the ability of the methods to learn complex physical dynamics, like the guiding center from gyrokinetic plasma physics. 
}

\keywords{Hamiltonian dynamics, structure preserving integrator, symplectic integrators, neural models, geometric machine learning, guiding center}


\pacs[MSC Classification]{%
    65P10, 
    68T07, 
    37M15 
}

\maketitle

\section{Introduction}
\label{sec:Intro}

With the rise in machine learning research, the connection between neural networks and differential equations—particularly ordinary differential equations (ODEs)—has become an increasingly active area of study, explored through three main application domains.
The first approach interprets certain neural networks, such as Residual Networks (ResNets), as discretizations of ODEs \citep{Chen.Rubanova.ea.2018.NeuralOrdinaryDifferential,dupont2019augmented,massaroli2020dissecting}. This perspective not only enables the design of novel architectures but also provides deeper insight into the internal dynamics of neural networks.
The second line of research views gradient descent methods as discretizations of continuous-time gradient flows \citep{jacot2018neural}, allowing a more refined analysis of learning dynamics by studying trajectories in parameter space.
The third axis concerns time series prediction. Instead of using models such as Recurrent Neural Networks (RNNs) or Long short-term memory architectures
(LSTM), when it is reasonable to assume that the data stems from an underlying continuous process, it becomes relevant to model these series with ODEs or stochastic differential equations (SDEs) \citep{zhu2024dyngma}. This work belongs to that third category.

Many contributions have been made to the learning of differential equations: the aim is then to determine the vector field, whose integral lines fit the observed data. Some, such as Sparse Identification of Non-linear Dynamics (SINDy) \citep{brunton2016discovering,kaiser2018sparse}, aim to infer the analytic form of the governing equations from time transitions between pairs of states. Similar strategies have been developed using neural networks. Other works are based on unrolled training over multiple time steps \citep{lee2022structure}, either to learn discrete-time flows that are then composed, or to model continuous-time dynamics directly \citep{bilovs2021neural}.
In physical applications, learning ODEs is particularly relevant for model reduction \citep{farenga2025latent, bonneville2024comprehensive,regazzoni2023latent,cote2023hamiltonian,franck2025reduced} to capture the dynamics of latent variables or even to discover new physical laws \citep{tenachi2023deep}. 
From both the standpoint of physical modeling and long-term numerical stability, the preservation of conservative and dissipative structures is of key importance. This has naturally raised the question of whether such structures can be preserved by learning algorithms.

In the case of dissipative systems, some works have focused on the joint learning of a Lyapunov function and its associated ODE \citep{kolter2019learning}, or on separating the dissipitive and conservative parts~\citep{eidnes2023pseudo,Eidnes.Lye.2024.PseudoHamiltonianNeuralNetworks}. A large part of the literature, however, has focused on conservative ODEs, particularly those arising from Lagrangian or Hamiltonian formalisms. For such systems, it is well known that preserving the underlying geometric structures, especially symplectic structures, is crucial to ensure long-term stability.
In the context of machine learning, the seminal work on Hamiltonian Neural Networks (HNN) \citep{Greydanus.Dzamba.ea.2019.HamiltonianNeuralNetworks} proposed to learn the vector field as the gradient of a Hamiltonian function, described with a neural network. It demonstrated that preserving the Hamiltonian structure—i.e., directly learning the Hamiltonian function—significantly improves the long-term stability of the learned dynamics. 

\review{%
Since then, a range of approaches have been proposed to learn dynamics that respect Lagrangian \citep{Cranmer.Greydanus.ea.2020.LagrangianNeuralNetworks} or Hamiltonian structures \citep{Cranmer.Greydanus.ea.2020.LagrangianNeuralNetworks, David.Mehats.2021.SymplecticLearningHamiltonian, Lishkova.Scherer.ea.2022.DiscreteLagrangianNeural}, often from solution snapshots at discrete time-steps.
They rely either on structure-preserving mappings with SympNets~\citep{Jin.Zhang.ea.2020.SympNetsIntrinsicStructurepreserving,tapley2024symplectic} or HénonNets~\citep{Burby.Tang.ea.2021.FastNeuralPoincare,Duruisseaux.Burby.ea.2023.ApproximationNearlyperiodicSymplectica}, while others rely on well-chosen numerical schemes~\citep{David.Mehats.2021.SymplecticLearningHamiltonian} such as variational integrators~\citep{Ober-Blobaum.Offen.2023.VariationalLearningEuler,Offen.Ober-Blobaum.2024.LearningDiscreteModels}.
This latter approach is similar to deep solvers or HyperSolvers~\citep{Poli.Massaroli.ea.2020.HypersolversFastContinuousDepth,Shen.Cheng.ea.2020.DeepEulerMethod,Bouchereau.Chartier.ea.2023.MachineLearningMethodsa}, and may enable the learning of physical properties such as symmetries using backwards error analysis~\citep{Lishkova.Scherer.ea.2022.DiscreteLagrangianNeural}. Sometimes both approaches may be combined~\citep{Mathiesen.Yang.ea.2022.HyperverletSymplecticHypersolver}.
}

Most of these contributions focus on canonical Hamiltonian systems, where the dynamics are expressed in $\mathbb{R}^{2d}$ with the standard symplectic form. However, many physical systems (e.g., fluid mechanics, plasma physics, or constrained mechanical systems) do not follow this canonical form. This leads to non-canonical systems, where the underlying geometry is more complex and often described by generalized Poisson structures. Such dynamics are then described by both a Hamiltonian function and a skew-symmetric space-dependent matrix, defining the underlying Poisson structure: this matrix is invertible for non-canonical Hamiltonian system and non-invertible for general Poisson system. 
\review{In model order reduction of dynamical systems, allowing for noncanonical representations can be advantageous over restricting to canonical representations, as noncanonical formulations often allow for lower-dimensional solution spaces and/or extended expressivity.}
A few recent works have begun to extend previous methods to these non-canonical cases \citep{Chen.Matsubara.ea.2021.NeuralSymplecticForm} and to Poisson systems \citep{Jin.Zhang.ea.2020.LearningPoissonSystems, Burby.Tang.ea.2021.FastNeuralPoincare}.

\begin{color}{Review}

\subsection{Setting}

We are interested in performing long-time simulations from neural non-canonical Hamiltonian models, building on the work of \cite{Chen.Matsubara.ea.2021.NeuralSymplecticForm} which extended the seminal work of \cite{Greydanus.Dzamba.ea.2019.HamiltonianNeuralNetworks} to the non-canonical setting.
In addition to learning a Hamiltonian (i.e. a scalar potential), this approach learns a symplectic potential (a vector potential) from which the non-canonical structure is derived.%

Structure preservation is key for medium-term accuracy, but long-term predictions require the additional use of an appropriate numerical integrator.
This is also the case for known (non-learnt) dynamics, see e.g. \cite{Hairer.Lubich.ea.2006.GeometricNumericalIntegration}:
the energy dissipation from standard integration schemes (such as RK4) creates an inaccurate qualitative behaviour of the solution over a long time.

This paper focuses on the use of a time-stepping scheme appropriate for long-time simulation on the learnt dynamics from two different learning strategies, either from vector-field data or from phase-space snapshots, as illustrated in Figure~\ref{fig:dataset_strategies}.

\bmhead{Time-stepping scheme}
We investigate the use of a recent numerical scheme by~\cite{Ellison.Finn.ea.2018.DegenerateVariationalIntegrators} on the learnt dynamics. 
This so-called Degenerate Variational Integrator (DVI) has been shown to produce highly reliable numerical solutions for non-canonical problems over long times in the context of plasma physics. 
However, it requires a strong assumption on the problem, namely that half of the coordinates of the symplectic potential are zero.

\begin{figure}
    \centering
    \includegraphics[width=.49\textwidth]{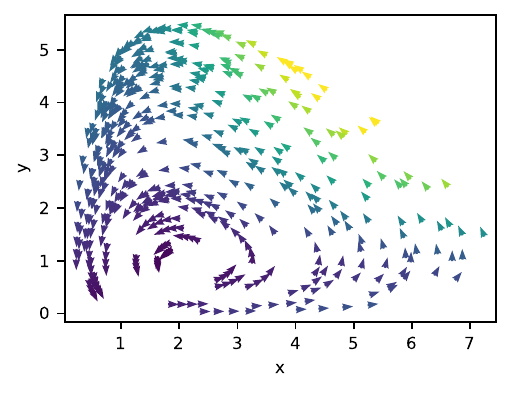}\hfill%
    \includegraphics[width=.49\textwidth]{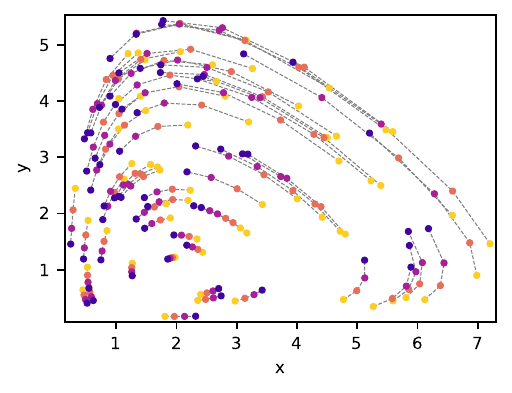}%
    \vspace*{-.2cm}%
    \caption{\review{\emph{The different datasets for the two learning strategies} for the Lotka-Volterra problem (\S\ref{subsec:long_time_intro}), extracted from the validation set. Left: vector-field learning, color indicates magnitude. Right: scheme learning, color indicates time evolution.}}
    \label{fig:dataset_strategies}
\end{figure}

\bmhead{Vector-field learning} 
When a reference model is known, then the neural vector-field can be fitted with standard supervised learning.
Given collocation points $(z^{(i)}, \dot z^{(i)})_{i \in \mathcal{I}}$ where $\mathcal{I} \subset \mathbb{N}$ indexes the data, the velocities $\dot z^{(i)}$ give the values of the vector-field at these points.

\bmhead{Scheme learning}
The vector field is not known at any collocation points but snapshots along trajectories are known.
In that case, the dataset consists of pairs $(z_{|t = 0}, z_{| t = h}) \in M^2$ with some time-step $h$.
This is the original setting of Neural ODEs~\citep{Chen.Rubanova.ea.2018.NeuralOrdinaryDifferential}, where the authors fit a (learnt) vector field such that trajectories along it map the input to the output.
It may also be used with reduced order modeling to learn unknown dynamical systems from the reduced data at discrete time-steps.

\subsection{Contribution}

When reproducing results from geometric learning~\citep{Chen.Matsubara.ea.2021.NeuralSymplecticForm} and numerical methods~\citep{Ellison.Finn.ea.2018.DegenerateVariationalIntegrators}, we show that combining the two may require some specific consideration.

Using the DVI to simulate a neural non-canonical problem, we observe that the scheme sometimes produces a completely incorrect prediction. 
In general, this is a reminder that neural models present unexpected stiffness compared to analytical ones.
In particular, this stems here from the fact that the scheme does not only simulate the evolution in the phase space, it also involves the evolution of the symplectic potential---it is not \emph{gauge invariant}.
This means that its error bound may increase if it is not penalised in the learning process, rendering it useless for simulating the learnt problem even over short times.
To our knowledge, this gauge-dependency of the DVI was not previously identified or studied.

Once the neural models stabilised, we compare the results of vector-field learning (VF learning) and scheme learning.
This is done on several test cases: a standard Lotka-Volterra model, a massless charged particle, and the guiding center, a 4-dimensional reduced model from gyrokinetics in plasma physics.
Throughout the paper, we aim for a pedagogical presentation to introduce these uncommon tools and problems.

Our contributions may be summarized as follows:

\bmhead{Enabling the DVI for neural models}
We perform an error analysis of the 1st-order DVI from which we define a penalization term to make VF learning compatible with the scheme.
For scheme learning, we avoid trivial solutions and later introduce a simpler loss function for pre-training on larger problems.

\bmhead{Comparison of learning strategies} 
Most papers implement only one of the two training types. By using both, we illustrate which applications may benefit or suffer from either approach.

\bmhead{Multi-scale error}
The guiding center model has components with very different magnitudes.
However, the nature of the (geometric) architectures makes a scaling the outputs of the neural network irrelevant.
Therefore, we introduce a data-informed loss based on a Gram matrix to treat each component the same.

\end{color}

\subsection{Outline}

The outline of this paper is as follows. Section~\ref{sec:ODEs} presents non-canonical Hamiltonian framework and the DVI numerical scheme, based on the Lagrangian formulation. Numerical tests also illustrate the advantages of the DVI for long-time simulations and issues related to gauge invariance when used in geometric model learning. In Section~\ref{sec:learnVF} and Section~\ref{sec:learnNum}, we introduce the vector-field learning and the scheme learning strategies, adapted to or based on to DVI. Finally, a series of numerical experiments (Lotka-Volterra, massless charged particle and guiding center) is conducted in Section~\ref{sec:simulations} to assess the ability of both methods to capture the dynamics.

\section{Non-canonical Hamiltonian dynamics}
\label{sec:ODEs}

Canonical Hamiltonian systems form a fundamental class of dynamical systems with a specific geometric structure. They are written as
\[
\dot{z} = J^{-1} \nabla H(z),
\]
where $z(t) \in \R^{2d}$ denotes the time-dependent coordinates, $H(z) \in \R$ denotes the Hamiltonian function representing the total energy of the system, and \( J \) is the canonical symplectic matrix given by
\[
J = \begin{bmatrix} 0 & -\mathrm{Id} \\ \mathrm{Id} & 0 \end{bmatrix} \in M_{2d}(\R).
\]
These systems have been extensively studied from theoretical and numerical perspectives, and more recently from a machine learning point of view.
However, many physically relevant problems are not naturally expressed in canonical coordinates, even though they retain an underlying Hamiltonian structure. These systems, known as non-canonical Hamiltonian systems, take the more general form:
\begin{equation} \label{eq:ode_z}
    \dot{z} = W(z)^{-1} \nabla H(z) ,
\end{equation}
where $W(z) \in M_{2d}(\R)$, skew-symmetric, defines a \textit{closed} and non-singular 2-form (i.e. its inverse defines a Poisson bracket).

\subsection{A subclass of non-canonical problems}
\label{subsec:non-canonical}

We will focus on a subclass of non-canonical problems, where the symplectic form $W$ is defined as the \textit{exterior derivative} of a one-form, which guarantees its closedness.\footnote{%
\review{In other words, we assume that the symplectic form $W$ is \emph{exact} instead of \emph{closed}.}%
} It means that it is obtained from a mapping $A : \mathbb{R}^{2d} \rightarrow \mathbb{R}^{2d}$, called the symplectic potential, through the relation
\begin{equation}
    \label{eq:closed_W}
    W = -{\sf d} A = (D_z A)^\mathsf{T} - D_z A ,
\end{equation}
where $D_z A$ denotes the Jacobian of $A$. In terms of coordinates, $(D_z A)_{ij} = \frac{\partial A_i}{\partial z^j}$.

The symplectic form must derive from a potential for two reasons.
Firstly, \cite{Chen.Matsubara.ea.2021.NeuralSymplecticForm} observed that learning the symplectic form as a skew-symmetric matrix without additional structure generates neural models that are often unstable, with rapidly exploding solutions.\footnote{%
    This might be due to the fact that the symplectic potential ensures the existence of Poincaré invariants (similarly to the Hamiltonian ensuring the existence of an energy invariant).
    Similarly, the behavior of geometric methods relies on generating functions~\citep[VI.5]{Hairer.Lubich.ea.2006.GeometricNumericalIntegration}, which use the symplectic potential.
} 
Secondly, we later use a numerical method (described next in \S\ref{sec:dvi}) which requires that the symplectic potential takes the following simplified form: 
\begin{equation}
    A(z) = \begin{bmatrix} \vartheta(z) \\ 0 \end{bmatrix}.
    \label{eq:potentialA}
\end{equation} 
With this choice, denoting  $z = (x, y)^\mathsf{T}$ with $x, y \in \mathbb{R}^d$, the symplectic two-form is the block matrix 
\begin{equation}
    \label{eq:symplecticForm}
    W = \begin{bmatrix}
        (D_x \vartheta)^\mathsf{T} - (D_x \vartheta) & -(D_y \vartheta) \\
        (D_y \vartheta)^\mathsf{T} & 0
    \end{bmatrix} ,
\end{equation}
with $D_x\vartheta$ and $D_y\vartheta$ denoting the Jacobian of $\vartheta$ w.r.t. $x$ and $y$ respectively as in~\eqref{eq:closed_W}.
Then system~\eqref{eq:ode_z} writes:
\begin{align*}
    &(D_y \vartheta)^\mathsf{T}\ \dot x = \nabla_y H ,\\
    &\bigl( (D_x\vartheta) - (D_x \vartheta)^\mathsf{T}\bigr)\, \dot x + (D_y \vartheta)\: \dot y = -\nabla_x H ,
\end{align*}
where the gradient is the transpose of the differential.

\begin{remark}
    Canonical Hamiltonian systems belong to this category of problems, setting $\vartheta(x, y) = y$. 
    Because every non-canonical system admits a local change of variable 
    which transforms it into a canonical problem, this ansatz on the one-form may always be satisfied, at least locally, up to a change of variable.
    \review{Without a change of variable, this split between $x$ and $y$ might not always be possible, while the global existence of a symplectic potential $z \mapsto A(z)$ is guaranteed as long as the domain or symplectic form respectively have no hole or pole (see e.g.~\cite{Chen.Matsubara.ea.2021.NeuralSymplecticForm} for some details).}
\end{remark}

This has the immediate advantage of characterising the invertibility of $W$ by the local injectivity of $y \mapsto \vartheta(x,y)$ at fixed $x$, giving a direct criterion for well-posedness of the problem. 
The system may also readily be inverted, yielding the vector field
\begin{equation}
    \label{eq:degenerate_vf}
    \begin{cases}
        \dot x = (D_y \vartheta)^{-\mathsf{T}} \nabla_y H , \\
        \dot y = -(D_y\vartheta)^{-1} \left( (D_x\vartheta - D_x\vartheta^\mathsf{T}) (D_y \vartheta)^{-\mathsf{T}} \nabla_y H + \nabla_x H \right) .
    \end{cases}
\end{equation}
From this identity, one may apply standard numerical methods, as did~\cite{Chen.Matsubara.ea.2021.NeuralSymplecticForm} where high-order Runge-Kutta schemes with adaptive time-steps are used. 
Here, however, we are interested in schemes specifically designed for this type of problems, namely ``degenerate variational integrators''.

\subsection{Degenerate variational integrators}
\label{sec:dvi}

Just as symplectic integrators are essential for the long-term numerical integration of canonical Hamiltonian systems \citep{Hairer.Lubich.ea.2006.GeometricNumericalIntegration},preserving the geometric structure and qualitative behavior of the exact flow—it is equally important to design structure-preserving integrators for non-canonical Hamiltonian systems. These systems often possess a generalized symplectic or Poisson structure that must be respected to ensure physically meaningful simulations over long time intervals.
One of the main approaches to achieving this goal is to recast the problem in a Lagrangian framework, and to construct discrete integrators that mimic the variational derivation of the equations of motion. This leads to the class of \emph{degenerate variational integrators} (DVIs) from \cite{Ellison.Finn.ea.2018.DegenerateVariationalIntegrators}, which extend the ideas of variational integration to degenerate or non-canonical settings, allowing for the preservation of geometric invariants even in the absence of a constant symplectic form. We now aim to recall this Lagrangian formalism and the construction of the corresponding scheme.

\bmhead{Degenerate Lagrangian dynamics} System \eqref{eq:ode_z}-\eqref{eq:closed_W} can be obtained by first defining the Lagrangian
\begin{equation}
    \label{eq:degenLagrangian}
    L(z, z_t) = A(z) \cdot z_t - H(z) ,
\end{equation}
and the dynamics is then recovered by extremizing the action $$\int_{t_0}^{t_1} L \left( z(t), \frac{\mathrm{d}}{\mathrm{d}t} z(t) \right) {\rm d}t,$$ yielding the so-called Euler-Lagrange equations: 
\begin{equation*}
    \frac{\rm d}{{\rm d} t} \left[ \frac{\partial L}{\partial z_t^i} \bigl( z(t), \dot z(t) \bigr) \right] = \frac{\partial L}{\partial z^i} \bigl( z(t), \dot z(t) \bigr),\quad \text{ for }i \in \{1, ..., 2d\},
\end{equation*}
which, owing to the chain rule, can be written as 
\begin{equation}\sum_{j=1}^{2d} \left[ \frac{\partial^2 L}{\partial z_t^i \partial z^j}\bigl( z(t), \dot z(t) \bigr) \dot z^j + \frac{\partial^2 L}{\partial z_t^i \partial z_t^j}\bigl( z(t), \dot z(t) \bigr) \ddot{z}^j \right] = \frac{\partial L}{\partial z^i}\bigl( z(t), \dot z(t) \bigr) .
\label{eq:eulerlagrange}
\end{equation}
This particular choice of Lagrangian is called \textit{degenerate}, as the vanishing of the Hessian w.r.t. $z_t$ removes the second-order derivative $\ddot{z}$ from this last equation.  Instead, it results in~\eqref{eq:ode_z} with the two-form~\eqref{eq:closed_W}. 

When considering a symplectic potential $A$ as in Eq. \eqref{eq:potentialA}, denoting  $z = (x, y)^{\mathsf{T}}$ with $x, y \in \mathbb{R}^d$, the Lagrangian writes
\begin{equation}
    \label{eq:Lagrangian}
     L (x, y, x_t, y_t) = \vartheta (x, y) \cdot x_t - H(x,y).
\end{equation}
To avoid any confusion with~\eqref{eq:degenLagrangian}, in the sequel we call this a \textit{properly degenerate} Lagrangian. We note that this arises naturally \review{in the canonical case}, using the so-called ``tautological one-form'' $\vartheta(q, p) = p$.

\begin{remark}
\review{%
    \cite{Burby.Finn.ea.2022.ImprovedAccuracyDegenerate} offer a slightly different notion of ``properly degenerate'' which amounts to assuming that the images of $D_z A$ and $D_z A^\mathsf{T}$ are complementary (therefore each is of dimension $d$).
    This condition is seemingly less restrictive than the $(x,y)$-split ansatz, but is impractical for machine learning and numerical simulations, therefore we will only consider~\eqref{eq:Lagrangian} instead of~\eqref{eq:degenLagrangian}.
}
\end{remark}

\bmhead{Degenerate variational integrators (DVI)}
Variational integrators are numerical schemes based on a discrete Lagrangian $(z_0, z_1) \mapsto L_h(z_0, z_1)$, for instance $L_h(z_0, z_1) = L \bigl( z_0, (z_1 - z_0)/h \bigr)$. 
The time-stepping scheme is then obtained from the \textit{discrete} Euler-Lagrange equations, defined as $$\nabla_{z_n} [L_h(z_{n-1}, z_n) + L_h(z_n, z_{n+1})] = 0.$$

Here we consider the \textit{degenerate} variational integrator from~\cite{Ellison.Finn.ea.2018.DegenerateVariationalIntegrators} which relies on the ansatz~\eqref{eq:Lagrangian}, and is derived from the discrete Lagrangian
\begin{equation}\label{eq:discrete-degenerate-lagrangian}
    L_h(x_n, x_{n+1}, y_{n+1}) = \vartheta (x_{n+1}, y_{n+1}) \cdot \frac{x_{n+1} - x_n}{h} - H(x_{n+1}, y_{n+1}) . 
\end{equation}
The time-stepping scheme stems from the discrete Euler-Lagrange equations, which in this case are written
\begin{equation*}
    \begin{cases}\displaystyle
        \nabla_{x_n} \left[ L_h(x_{n-1}, x_n, y_n) + L_h(x_n, x_{n+1}, y_{n+1}) \right] = 0, &\\
    	\nabla_{y_{n+1}} \left[ L_h(x_n, x_{n+1}, y_{n+1}) + L_h(x_{n+1}, x_{n+2}, y_{n+2}) \right] = 0,\quad &
    \end{cases}
\end{equation*}
Usually, this would be written differentiating w.r.t. $y_n$ and not $y_{n+1}$, but since these equations have to be satisfied for every $n$, the identity above is equivalent to the usual one.
This generates the time-stepping scheme
\begin{equation}
    \label{eq:Scheme}
    \begin{cases}
        \vartheta (x_{n+1}, y_{n+1}) = \vartheta (x_n, y_n) + (D_x\vartheta(x_n, y_n))^\mathsf{T}(x_n - x_{n-1}) - h \nabla_x H(x_n, y_n) , \\
        (D_y\vartheta(x_{n+1}, y_{n+1}))^\mathsf{T} x_{n+1} = (D_y\vartheta(x_{n+1}, y_{n+1}))^\mathsf{T} x_n + h \nabla_y H(x_{n+1}, y_{n+1}) .
    \end{cases}
\end{equation}
While this may seem like a two-step method with the term $x_{n-1}$ in the first equation, this term may be obtained explicitly using the second equation,
\begin{equation*}
     x_{n-1} = x_n - h \bigl( D_y \vartheta (x_n, y_n) \bigr)^{-\mathsf{T}} \nabla_y H(x_n, y_n) .
\end{equation*}
As with the continuous problem, the scheme is well-defined if $D_y \vartheta$ is invertible (up to stability and regularity conditions).
In practice, only $x_{-1}$ is computed. For later time-steps, the term $x_{n-1}$ can be used from previous iterations.

We implemented this scheme using the automatic differentiation (AD) routines of \texttt{PyTorch} to derive the Euler-Lagrange equations.\footnote{%
    It should be noted that $x_{n-1}$ and $x_{n+1}$ are assumed \textit{independent} of $x_n$ when deriving the Euler-Lagrange equations. In \texttt{PyTorch} vernacular, they must be ``detached'' from $x_n$ before differentiating.
}
Because we only simulate problems in low dimensions, Equation \eqref{eq:Scheme} is solved using a Newton method---the Jacobian also being computed using AD.
The initial guess of $(x_{n+1}, y_{n+1})$ in the Newton iterations is chosen using a single step of the explicit Euler method applied to~\eqref{eq:degenerate_vf}.

\begin{remark}\label{rem:gauge}
    The variational integrator is sensitive to gauge choice for the potential $\vartheta$. Indeed, changing $\vartheta(x,y)$ into $\vartheta(x,y) + g(x)$, with a function $g(x)$ with symmetric Jacobian,
    \begin{equation*}
        D_x g(x)^\mathsf{T} - D_x g(x) = 0,
    \end{equation*}
    does not modify the symplectic matrix $W$ (see Eq. \eqref{eq:symplecticForm}) and thus System \eqref{eq:degenerate_vf}, but change the approximate solution computed with \eqref{eq:Scheme}. 
    Although the convergence holds whatever the gauge is, the error constant may become quite large. 
    Numerical examples will be provided in Section \ref{sec:simulations}.
\end{remark}


\begin{color}{Review}
\bmhead{Non-uniform time-stepping}
It is well-known that using variable time-steps on geometric problems requires some specific considerations.
With Lagrangian systems, this is often done by introducing a new canonical pair $(s, \xi)$ such that the \emph{extended} Lagrangian
\begin{equation*}
    L_\mathrm{ext}(x,s,y,\xi, x_t, s_t) = \vartheta(x, y)\cdot x_t + \xi s_t - \bigl( H(x, y) + \xi \bigr) 
\end{equation*}
generates the same dynamics on $t \mapsto \bigl(x(t), y(t) \bigr)$ while additionally tracking the time evolution with the canonical part $\dot s = 1, \dot \xi = 0$.
This allows a time rescaling $\mathrm{d}\tau = \rho(x, y) \mathrm{d}t$, using constant time-steps on $\tau$ while recovering the true time-dependency with $s(\tau) = t$ so long as $s(\tau_0) = t_0$.
Indeed, denoting $Q = (x, s, y, \xi)$, the extended action satisfies
\begin{equation*}
\int_{t_0}^{t_1} L_\mathrm{ext} \left( Q(t), \frac{\mathrm{d}}{\mathrm{d}t} Q(t) \right) {\rm d}t = \int_{\tau_0}^{\tau_1} L_\mathrm{ext}\left( Q(\tau), \rho(z) \frac{\mathrm{d}}{\mathrm{d}\tau} Q(\tau) \right) \frac{1}{\rho(z)} \mathrm{d}\tau .
\end{equation*}
The symplectic potential which corresponds to the linear part in $\frac{\mathrm{d} Q}{\mathrm{d}\tau}$ is unchanged, but the Hamiltonian is now $(x, s, y, \xi) \mapsto \frac{H(x, y) + \xi}{\rho(x, y)}$.
Symbolically replacing $t$ by $\tau$ in the Euler-Lagrange equations using this new Hamiltonian, we find the dynamics involving $\frac{\mathrm{d}Q}{\mathrm{d}\tau}$, from which the numerical method can be applied.
%
\end{color}

\subsection{Long time simulation}
\label{subsec:long_time_intro}

As discussed in this section, long time simulations using neural models requires both geometric numerical integration and structure-preserving learning strategies. 
We first compare the long-time behavior of the standard RK4 scheme with the DVI (\S\ref{sec:dvi}), justifying the use of this specific scheme.
Then, by partially reproducing the study of~\cite{Chen.Matsubara.ea.2021.NeuralSymplecticForm}, we illustrate the importance of preserving the underlying geometric structure during training, even without numerical considerations.
Finally, we show that simply applying the DVI to a neural model may result in unstable solutions, prompting the development of specific training strategies in the next section.

\bmhead{Reference model}
This study is conducted on a classical Lotka-Volterra problem which can be found in~\cite{Hairer.Lubich.ea.2006.GeometricNumericalIntegration}, 
\begin{equation}
\label{eq:LotkaVolterraDiffEq}
    \dot x = x(1 - y) , \qquad
    \dot y = y(x - 2) . 
\end{equation}
This corresponds to the structure \eqref{eq:degenerate_vf} with maps
\begin{equation}
    \label{eq:LotkaVolterraLagrangian}
    \vartheta(x,y) = -\ln(y) / x, \qquad 
    H(x,y) = x + y - 2\ln(x) - \ln(y) .
\end{equation}
In this case, a \review{global} change of variable $(q,p) = \bigl( \ln(x), \ln(y) \bigr)$ yields a canonical structure with Hamiltonian $(q,p) \mapsto e^q + e^p - 2 q - p$. \review{From this, high-order geometric numerical methods could be applied to ensure long-time accuracy and stability.} In general, however, such a change of variable is defined locally, not globally.  We shall therefore treat this problem as purely non-canonical \review{in the sequel}.

\begin{figure}
    \centering%
    \includegraphics[width=.49\textwidth]{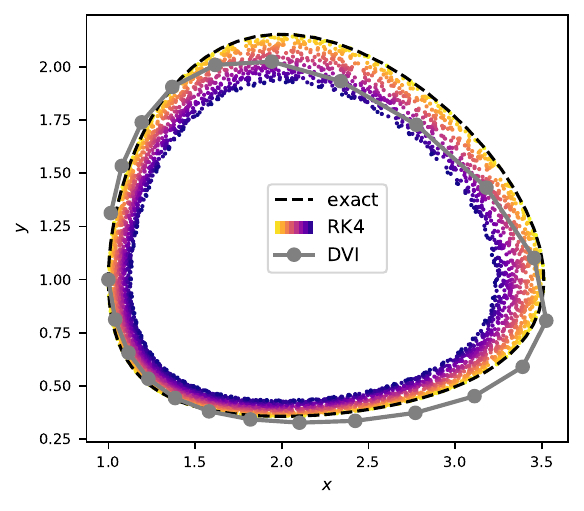}\hfill%
    \includegraphics[width=.49\textwidth]{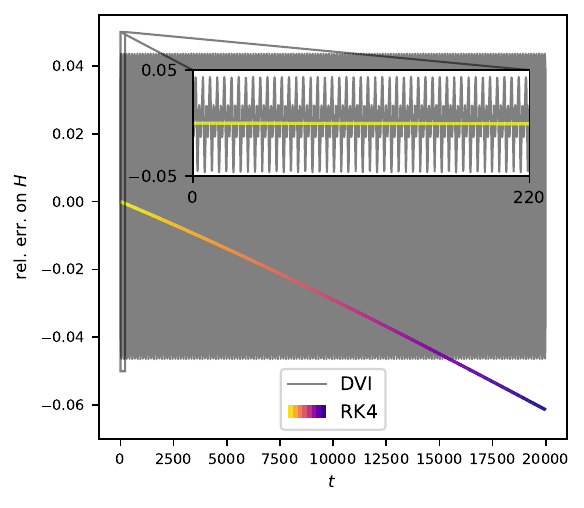}%
    \caption{(Lotka-Volterra) \emph{Long-time \review{energy dissipation of the RK4 scheme compared to the DVI}}, with initial condition $x_0 = 1$, $y_0 = 1$ and time-step $h = 0.2$. Left: solutions in phase space. Only one out of 16 time-steps is displayed for the RK4 solution (to reduce file size), and only the first time-steps of the DVI solution is displayed (the solution is periodic). Right: relative error on the Hamiltonian as a function of time.}
    \label{fig:refLotkaVolterra}
\end{figure}

\bmhead{Using the DVI}
Another important aspect of long-time simulations mentioned previously is the numerical scheme, which must be appropriate---we introduced the DVI to that effect.
Figure~\ref{fig:refLotkaVolterra} compares the results of long-time simulations on the reference model~\eqref{eq:LotkaVolterraDiffEq} using either the standard RK4 (explicit Runge-Kutta of order 4) or the DVI with a large time-step.
In short time (clear yellow dots), the RK4 scheme is highly-accurate, as expected.
In long time, however, the numerical solution does not stay on the closed orbit and becomes less and less accurate.
Here the scheme dissipates energy
as can be seen on the right-hand plot.
In contrast, the DVI \textit{does} generate a periodic solution, though its orbit is clearly inaccurate (as expected from a first-order scheme).
This means that the error is fairly large in short time, but the preservation of qualitative properties ensures that the long-time behaviour of the numerical solution is closer to that of the exact solution.\footnote{By exact solution, we mean a highly-accurate numerical solution computed using the \texttt{solve\_ivp} routine of SciPy with parameters \texttt{atol=1e-12} and \texttt{rtol=1e-10}.}

\bmhead{Defining neural networks}
In \cite{Chen.Matsubara.ea.2021.NeuralSymplecticForm}, the authors enforce the symplectic structure by learning both the Hamiltonian $H_\Theta: \R^{2d} \to \R$ and the symplectic potential $A_\Theta: \R^{2d} \to \R^{2d}$.
Echoing the presentation of \S\ref{subsec:non-canonical}, this allows the construction of the symplectic form $z \mapsto W_\Theta(z) \in M_{2d}(\R)$ with automatic differentiation (AD) using~\eqref{eq:closed_W}, to find the vector field 
\begin{equation}
\label{eq:structNeuralNet}
    f_\Theta^{\text{non-can}}(z) := W_\Theta(z)^{-1} \nabla H_\Theta(z) 
\end{equation}
(where $\nabla H$ is also computed using AD).
The networks $A_\Theta, H_\Theta$ are fitted on collocation points $(z^{(i)}, \dot z^{(i)})_{i \in \mathcal{I}}$ where $\mathcal{I} \subset \mathbb{N}$ indexes the data and $\dot z^{(i)} = f(z^{(i)})$.
This is done by minimizing the loss 
\begin{equation}
\label{eq:fitCMY}
    \mathcal{L}_\mathrm{vf}(\Theta) := \frac{1}{|\mathcal{I}|} \sum_{i \in \mathcal{I}} \| \dot z^{(i)} - f_\Theta(z^{(i)}) \|^2 , 
\end{equation}
using an optimization method in batches.

Here we further impose the ansatz \eqref{eq:potentialA} nullifying the second half of the coordinates of $A_\Theta$, and therefore learn only the first half of the symplectic potential $\vartheta_\Theta: \R^{2d} \to \R^d$.
This allows the use of the Degenerate Variational Integrator \citep{Ellison.Finn.ea.2018.DegenerateVariationalIntegrators} presented in~Section \ref{sec:dvi}.
To recover the vector-field, the symplectic form is constructed with~\eqref{eq:symplecticForm}.

\bmhead{Hard-coding structure}
To show that the structure of model~\eqref{eq:structNeuralNet} is important, let us compare the solutions of three different neural models with the reference~\eqref{eq:LotkaVolterraDiffEq}. 
The baseline neural model is a single multi-layer perceptron (MLP) 
\begin{equation}
    z \mapsto f_\Theta^{\text{no-struct}}(z)
    \label{eq:nostructuremodel}
\end{equation}
(with 2 hidden layers of dimension 40) which represents the vector-field. 
The ``canonical'' neural model, usually called HNN~\citep{Greydanus.Dzamba.ea.2019.HamiltonianNeuralNetworks}, consists of a single MLP $H_\Theta$ from which the vector-field is obtained by automatic differentiation (AD) as 
\begin{equation}
    z \mapsto f_{\Theta}^{\text{can}}(z) := J^{-1} \nabla H_\Theta(z).
    \label{eq:canonicalmodel}
\end{equation}
The non-canonical model \eqref{eq:structNeuralNet}, consists of two separate MLPs, the symplectic potential $\vartheta_\Theta$ and the Hamiltonian $H_\Theta$, both of 2 hidden layers of size 30.
For all networks, two trainings are performed using the Adam optimizer and batches of size 500. The first training uses a learning rate $\mathrm{lr} = 10^{-2}$ for 50 epochs, and the second $\mathrm{lr} = 10^{-3}$ for 150 epochs.
The dataset and the MLP architecture will be detailed in the next sections.

\begin{figure}
    \centering%
    \includegraphics[width=.49\textwidth]{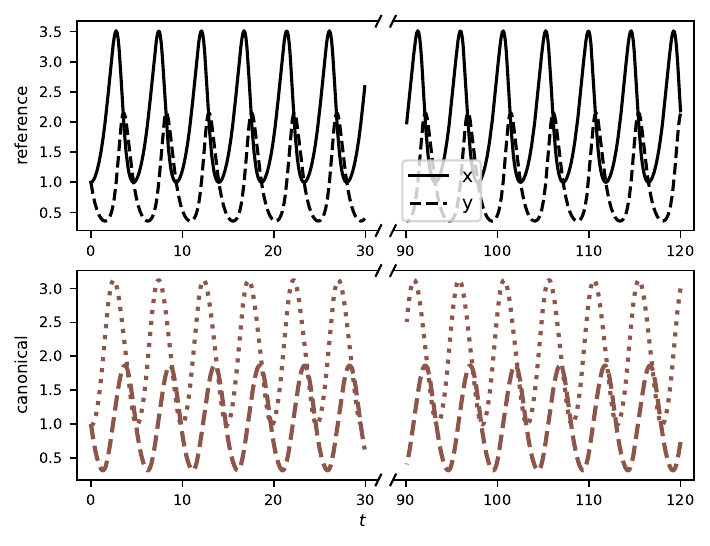}\hfill%
    \includegraphics[width=.49\textwidth]{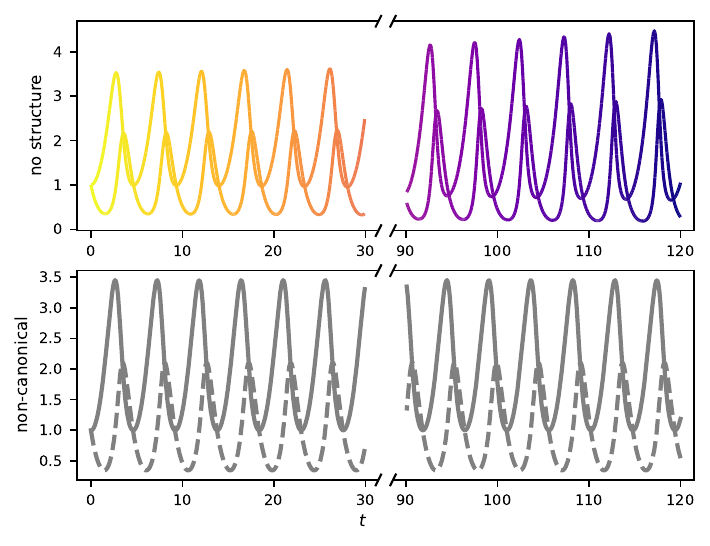}%
    \caption{(Lotka-Volterra) Solutions with initial condition $x_0 = 1$, $y_0 = 1$ for the different models (reference model \eqref{eq:LotkaVolterraDiffEq}, no structure neural model \eqref{eq:nostructuremodel}, canonical neural model \eqref{eq:canonicalmodel}, non-canonical neural model \eqref{eq:structNeuralNet}), obtained using \texttt{solve\_ivp}.}
    \label{fig:architecturesSolutionsLotkaVolterra}
\end{figure}

\begin{figure}
    \centering%
    \includegraphics[width=.49\textwidth]{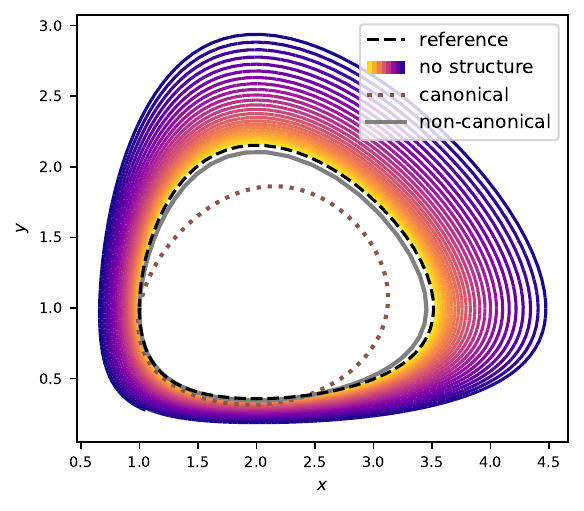}\hfill%
    \includegraphics[width=.49\textwidth]{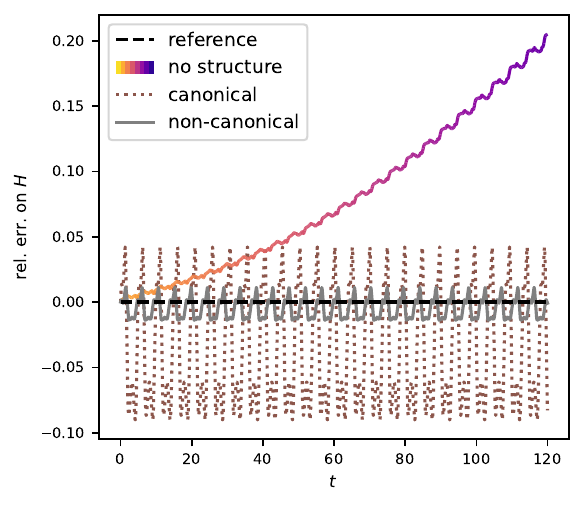}%
    \caption{(Lotka-Volterra) \review{\emph{Blow-up of the neural model without structure.}} Solutions with initial condition $x_0 = 1$, $y_0 = 1$ for the different models (reference model \eqref{eq:LotkaVolterraDiffEq}, no structure neural model \eqref{eq:nostructuremodel}, canonical neural model \eqref{eq:canonicalmodel}, non-canonical neural model \eqref{eq:structNeuralNet}), obtained using \texttt{solve\_ivp}. Left: phase portrait. Right: time evolution of the relative error of the Hamiltonian \eqref{eq:LotkaVolterraLagrangian} for the different solutions.}
    \label{fig:architecturesLotkaVolterra}
\end{figure}

Figure~\ref{fig:architecturesSolutionsLotkaVolterra} displays a highly accurate numerical solution of each model, computed using the \texttt{solve\_ivp} routine of SciPy with parameters \texttt{rtol=1e-10} and \texttt{atol=1e-12}. 
The same solutions are shown in phase-space in Figure~\ref{fig:architecturesLotkaVolterra} (left). 
Instead of yielding a closed (periodic) orbit, the baseline model diverges over time, making it unsuitable for long-time simulation.
While the canonical model generates a closed orbit, it is also visible that it cannot accurately recreate the non-canonical vector-field\review{, demonstrating that the symplectic potential is just as important as the Hamiltonian}.
Despite the short training and the small size of the networks, the non-canonical model recovers accurate dynamics.

On the right-hand graphs, it appears that on the trajectory of this model, the (reference) Hamiltonian is preserved up to $\sim$2\%, while the baseline without structure diverges. This is computed by evaluating the Hamiltonian function~\eqref{eq:LotkaVolterraLagrangian} at every time for each model's trajectory. We do not compare the learnt Hamiltonian functions with the reference one, though each geometric neural model preserves its own Hamiltonian.

\begin{figure}
    \centering
    \includegraphics[width=0.33\textwidth]{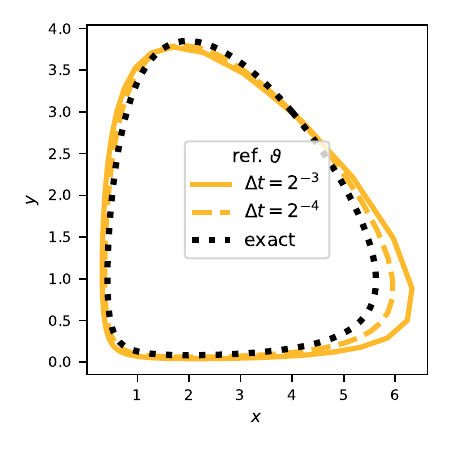}\hfill%
    \includegraphics[width=0.33\textwidth]{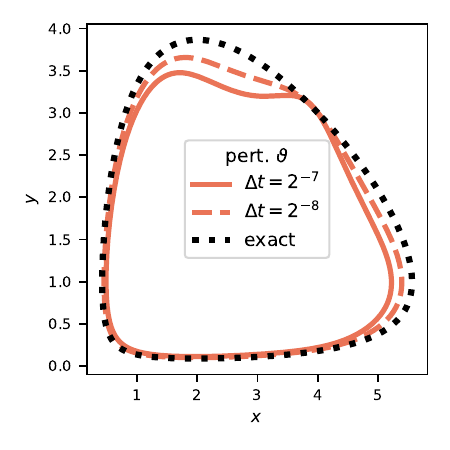}\hfill%
    \includegraphics[width=0.33\textwidth]{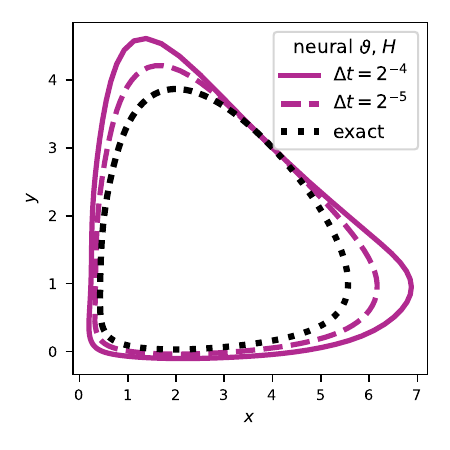}%
    \caption{(Lotka-Volterra)
        \review{\emph{Impact of the gauge perturbation on the DVI.}}
        solutions with initial condition $x_0 = 4$, $y_0 = 3$  obtained using the first-order DVI scheme with different time steps for different models. 
        Left: reference model~\eqref{eq:LotkaVolterraLagrangian}. 
        Middle: a perturbed model $\vartheta(x, y) \leftarrow \vartheta(x, y) + \frac12 \cos(2x)$. 
        Right: a non-canonical neural model \eqref{eq:structNeuralNet}. Exact solutions refers to refined solutions obtained using \texttt{solve\_ivp} on the same model.
    }
    \label{fig:perturbedLotkaVolterra}
\end{figure}

\bmhead{Unexpected instability}
We now show that the DVI is sensitive to the gauge of the model, which turn out to be important when the model is learnt from data.
Consider the rightmost plot of Figure~\ref{fig:perturbedLotkaVolterra}, which shows that the solution of the neural model with structure obtained as in~\cite{Chen.Matsubara.ea.2021.NeuralSymplecticForm} (dotted black) is accurate compared to the reference model (gray).
In purple is the numerical solution using the DVI for two different times steps.
Comparing it to the reference model (left plot), the DVI is apparently less stable for the neural model.
In fact, when plotting several trajectories, many diverge after a few iterations (this will be shown in \S\ref{subsec:LotkaVolterra}).

Since $d = 1$, the equations of the continuous dynamics~\eqref{eq:degenerate_vf} are simplified, as the antisymmetric part of $D_x \vartheta = \partial_x \vartheta$ simply vanishes:
\begin{equation*}
    \begin{cases}
        \dot{x} = \partial_y H (x,y) / \partial_y \vartheta (x,y), \\
        \dot{y} = -\partial_x H (x,y) / \partial_y \vartheta (x,y).
    \end{cases}
\end{equation*}
However, this simplification does not occur for the discrete scheme, which writes
\begin{equation}
    \begin{cases} \displaystyle
        \frac{\ln(y_{n+1})}{x_{n+1}} = \frac{\ln(y_n)}{x_n} \bigl(1 - h (1 - y_n) \bigr) + h\left( 1 - \frac{2}{x_n} \right), \\ \displaystyle
        x_{n+1} = x_n + h x_{n+1} ( 1 - y_{n+1} ) , 
    \end{cases}
\end{equation}
where we plugged in $x_n - x_{n-1} = h x_n (1 - y_n)$. 
While the second component is a standard implicit Euler method on $\dot x$, the first component is more exotic, involving a logarithm which would never appear when considering only the vector field.

The expression of the scheme shows that, while the continuous dynamics remain unchanged under any gauge perturbation of the potential (see Remark~\ref{rem:gauge}), this is not the case for the scheme. It thus becomes clear that if the learning process has captured a potential with a large perturbation depending only on $x$, then the scheme may produce completely different results.
This is illustrated in the center plot of Figure~\ref{fig:perturbedLotkaVolterra}, where we deliberately modify the potential to make the problem more stiff.
We argue that such gauge perturbations may occur during training, and propose a way to avoid them in the next section.

\section{Learning the vector field with regularization}
\label{sec:learnVF}

As previously discussed, the degenerate Lagrangian systems considered here are invariant under certain potential perturbations. Consequently, a learning procedure that preserves this structure may generate a whole family of potentials leading to the same trajectory. However, this property does not generally hold when using a structure-preserving scheme such as the Discrete Variational Integrator (DVI), which is not gauge-invariant and thus leads to the numerical issues discussed earlier (see Fig.~\ref{fig:perturbedLotkaVolterra}). 
Therefore, to apply a DVI to a learned non-canonical system, one must prevent the learning process from producing potentials that significantly deviate from the true one.
A fairly natural approach is to add a penalization term that constrains the space of admissible potentials. This can also be seen as introducing a prior on the potential being learned.

As in~\eqref{eq:fitCMY}, the networks $\vartheta_\Theta, H_\Theta$ are fitted on collocation points $(z^{(i)}, \dot z^{(i)})_{i \in \mathcal{I}}$ where $\mathcal{I} \subset \mathbb{N}$ indexes the data and $\dot z^{(i)} = f(z^{(i)})$ is given by the reference model.
The loss now also includes a regularization term $(z, \dot z) \mapsto r_\Theta(z, \dot z) \in \mathbb{R}^{2d}$ with weight $\varepsilon > 0$, and is therefore written 
\begin{equation}
\label{eq:LossVFErr}
    \mathcal{L}_{\rm vf}(\Theta) = \frac{1}{|\mathcal{I}|} \sum_{i \in \mathcal{I}} \Bigl[ \|\dot z^{(i)} - W_{\Theta}^{-1}(z^{(i)}) \nabla H_{\Theta}(z^{(i)})\|^2 + \varepsilon\, \| r_\Theta(z^{(i)}, \dot z^{(i)}) \|^2 \Bigr],
\end{equation}
As always, the optimization is performed in batches.

\begin{remark}
\review{%
    The fitted model may then use the adaptive time-step approach presented in~\S2.2.
}
\end{remark}

The key challenge now is to propose a penalization term that effectively prevents undesirable symplectic potentials.

\subsection{The regularization term}

When performing a standard (local) error analysis on the DVI (see \S\ref{appendix:error}), denoting $x_h$ the solution after one time-step, we find the dominant error term 
\begin{equation}
\begin{cases}\displaystyle
    x(h) - x_h = -\frac{h^2}{2} \ddot x + \mathcal{O}(h^3), \\ \displaystyle
    y(h) - y_h = \frac{h^2}{2} D_y\vartheta^{-1} \left( \ddot \vartheta + D_x\vartheta \ddot x \right) + \mathcal{O}(h^3) .
\end{cases}
\end{equation}
On one hand, this is unsurprising: the error of a scheme of order $p$ generally only involves time-derivatives of order $p+1$ and higher.
On the other, these time-derivatives are usually in the coordinates (here $x$ and $y$), whereas here the gauge dependency is clear: the solution is accurate only if $\vartheta$ is sufficiently smooth, the notion of sufficient depending on the order of the scheme used.

Since our final goal here is to apply the first-order DVI, the penalization term $(z, z_t) \mapsto r_\Theta(z, z_t)$ is chosen to be precisely this dominant error term (removing the value $h$ which can be arbitrary). 
\review{%
This is specific to the goal at hand, and is different from gauge fixing.
Here, the symplectic potential that is learnt depends on the domain on which the learning occurs: considering a specific trajectory or a larger domain may yield different results.
Using a different numerical methods (such as a second-order DVI) would generate a different numerical error, thus may also yield a different result.\footnote{%
\review{
    Generalizing this approach would require a rigorous study of other schemes in order to find a general pattern, or a precise definition of what it means for $t \mapsto \vartheta\bigl(x(t), y(t)\bigr)$ to be ``the smoothest''. Both are beyond the scope of this paper.
}}
Since our goal is only to learn a minimizing potential, there is no guarantee that it is unique (even on a given domain), and different trainings may yield different results.
}

In practice, this penalization term is computed by inverting the linear system
\begin{equation}
\label{eq:schemeErrorTerm}
    \begin{bmatrix}
        D_x\vartheta_\Theta(z) + \bigl(D_x\vartheta_\Theta(z) \bigr)^\mathsf{T} & D_y\vartheta_\Theta(z) \\
        \bigl( D_y\vartheta_\Theta(z) \bigr)^\mathsf{T} & 0
    \end{bmatrix} r_\Theta(z, z_t)
    = D_z [\nabla_z L_\Theta(z, z_t)] z_t ,
\end{equation}
where $L_\Theta(z, z_t) = \vartheta_\Theta(x, y)^\mathsf{T} x_t - H_\Theta(x, y)$, with the differentials only applied to the corresponding argument, not to $z_t$. 
The derivation of this expression is detailed in Appendix~\ref{appendix:error}.

As before, this cost function involves a number of derivatives and thus requires automatic differentiation to compute the derivatives of the Lagrangian and Hamiltonian, as well as to account for the derivative of the matrix inverse with respect to the potential parameters appearing in the matrix. Since the matrices are small, differentiating through the matrix inversion does not necessarily require additional treatment, such as applying the implicit function theorem \citep{blondel2024elements}.
Now that the modified learning problem has been introduced, we propose to provide the technical details of the model and those necessary for the learning process.

\subsection{Hyperparameters of neural networks}
\label{subsec:nns}

\bmhead{Parameterization} The parameterized symplectic potenial $\vartheta_{\Theta}$ and Hamiltonian $H_{\Theta}$ are both defined as standard multi-layer perceptron (MLP) neural networks, of the form 
\begin{equation}
    \label{eq:architectureMLP}
    z \mapsto \mathcal{F}_{\Theta_L} \circ \rho_{\mu_{L-1}} \circ \mathcal{F}_{\Theta_{L-1}} \circ ... \circ \rho_{\mu_1} \circ \mathcal{F}_{\Theta_1}(z) ,
\end{equation}
with parametric linear layers $\mathcal{F}_{\Theta_\ell}$ and activation layers $\rho_{\mu_\ell}$ and $$\Theta = (\Theta_L, \mu_{L-1}, \Theta_{L-1}, ..., \mu_1, \Theta_1).$$
The linear layers correspond to an affine mapping $\mathcal{F}_{\Theta_\ell}(h) = M_\ell h + b_\ell$ with parameters $\Theta_\ell = (M_\ell, b_\ell)$ a weight matrix and vector bias respectively. 
For activation layers, we chose the so-called ``self-scalable $\tanh$'' of~\cite{Gnanasambandam.Shen.ea.2023.Stanh},  $\rho_{\mu_\ell}(h) = \tanh_\odot(h) + \mu_\ell \odot h \odot \tanh_\odot(h)$ with a parameter vector $\mu_{\ell}$, where $\odot$ denotes the componentwise product (or Hadamard product) and $\tanh_\odot$ is the hyperbolic tangent applied componentwise. This activation function was originally developed for multiscale physics-informed neural networks where the network is fitted using its differentials, which we also do here.

\bmhead{Normalizing the inputs}

The different coordinates might involve different scales, both for the input $z$ and the (indirect) output $\dot z$. For the input, we simply add a coordinate-wise normalization layer before each neural network, such that each coordinate has values in $[0, 1]$, i.e. there is a pre-processing layer
\begin{equation}
    \begin{bmatrix}
        z^1 \\ \vdots \\ \\ z^{2d} 
    \end{bmatrix} \mapsto \begin{bmatrix}
        \frac{z^1 - z^1_\mathrm{min}}{z^1_\mathrm{max} - z^1_\mathrm{min}}
        \\ \vdots \\
        \frac{z^{2d} - z^{2d}_\mathrm{min}}{z^{2d}_\mathrm{max} - z^{2d}_\mathrm{min}}
    \end{bmatrix} , 
\end{equation}
the minima and maxima being determined \textit{a priori} from the training set.

\bmhead{Normalizing the outputs}
It is standard for the outputs of neural networks to be normalized to facilitate learning. Here, this is crucial for the dynamics on all components to be accurate. In the forthcoming numerical examples, this will be especially true for the guiding center test case, in which some components differ by up to 4 orders of magnitude (see Figure~\ref{fig:gc:no_gram_rz}).

Since the outputs of neural networks do not correspond to the vector field directly, they cannot be rescaled. The vector-field data also cannot be normalized, since it corresponds to a physical quantity and determines the dynamics.
Therefore, the normalization is done in the training loss \eqref{eq:LossVFErr}: the standard Euclidean norm is replaced with the data-informed norm
\begin{equation}
    \label{eq:normVFErr}
    \| u \|^2 = u^\mathsf{T} M^{-1} u ,
\end{equation}
where $M = \frac{1}{|\mathcal{I}|} \sum_{i \in \mathcal{I}} \dot z^{(i)} (\dot z^{(i)})^\mathsf{T}$ is a Gram matrix associated with the data.

Because the Gram matrix is symmetric positive definite,\footnote{To avoid the case where $M$ is only symmetric positive \textit{semi}-definite, we actually add a small offset $\varepsilon I_d$.} we compute this norm using the inverse its symmetric (positive) square root $M = M^{1/2} M^{1/2}$, taking the Euclidean norm of $v = M^{-1/2} \delta\dot z$. 
This is obtained from a singular value decomposition $M = U \Sigma U^{\sf T}$ with $\Sigma = \operatorname{diag}(\sigma_1^{\:2}, ..., \sigma_{2d}^{\:2})$ and $U^{\sf T}\, U = I_{2d}$ (owing to the symmetry of $M$), setting $M^{-1/2} = U \Sigma^{-1/2} U^{\sf T}$.
Here this is computed ``on the fly'' for each batch, but it could be computed prior to the training.
If this decomposition is too costly, one might prefer a componentwise scaling $M = \frac{1}{|\mathcal{I}|} \sum_{i \in \mathcal{I}} {\rm diag}\bigl( \dot z_1^{(i)}, ..., \dot z_{2d}^{(i)} \bigr)^{2}$, to which a low-rank correction might also be applied.

\section{Learning the time-discrete dynamics through the scheme}
\label{sec:learnNum}

The second learning strategy aims at fitting the symplectic potential and the Hamiltonian so as to minimize the one-step error of the numerical scheme on a trajectory. As such, the training data consists of collocation triples $(z_0^{(j)}, z_1^{(j)}, z_2^{(j)})_{j \in \mathcal{J}}$ indexed on $\mathcal{J} \subset \mathbb{N}$, which correspond to data at times $(t^{(j)}, t^{(j)}+h, t^{(j)}+2h)$ for a given time step $h > 0$. 
We look for a symplectic potential $\tilde \vartheta_{\Theta}$ and a Hamiltonian $\tilde H_{\Theta}$ that nullify the mean one-step error of the scheme, denoted $S_\Theta$:
\begin{multline}
\label{eq:oneStepError}
    S_\Theta(z_0,z_1,z_2; h) =\\ 
    \begin{bmatrix} 
        \tilde \vartheta_{\Theta} (x_{2}, y_{2}) - \left(\tilde \vartheta_{\Theta} (x_1, y_1) + \bigl( D_x\tilde \vartheta_{\Theta} (x_1, y_1) \bigr)^\mathsf{T} (x_1 - x_0) - h \nabla_x \tilde H_{\Theta}(x_1, y_1)\right) \\
        (D_y\tilde \vartheta_{\Theta}(x_{2}, y_{2}))^\mathsf{T} x_{2} - \left((D_y\tilde \vartheta_{ \Theta}(x_{2}, y_{2}))^\mathsf{T} x_1 + h \nabla_y \tilde H_{\Theta}(x_2, y_{2})\right)
    \end{bmatrix}.
\end{multline}
\review{The same considerations of (local) existence of the symplectic potential and of the $(x,y)$-split apply as in the continuous case, but} the learnt quantities may \review{now} not exactly match the physical ones but a ``modified'' symplectic potential and Hamiltonian. Consequently, simulations on the learnt dynamics will be more accurate. 

\begin{remark}
    While similar, this approach presents two key differences with those of Deep Solvers~\citep{Shen.Cheng.ea.2020.DeepEulerMethod, Bouchereau.Chartier.ea.2023.MachineLearningMethodsa} and HyperSolvers~\citep{Poli.Massaroli.ea.2020.HypersolversFastContinuousDepth}.
    First, the network does not learn a correction term, it learns the entire Lagrangian.
    Second, the time-step is considered fixed, it is not an input of the neural networks.
    This first difference makes the learning more difficult, but the second makes it more simple---overall, it is just different.
    \review{Including the time-step as a parameter of the networks would not impact the results, but would not enable adaptive time-stepping either (see~\S\ref{subsec:timesteps}) and is therefore not considered here.}
    As pointed out in~\cite{David.Mehats.2021.SymplecticLearningHamiltonian}, using a scheme which preserves structure theoretically allows the network to reach a near-zero learning error.
\end{remark}

\subsection{The loss of scheme-learning}

Naively minimizing the mean squared norm of this error would be unsuccessful, because the trivial symplectic potential and Hamiltonian, $\tilde \vartheta_{\Theta} = 0$ and $\tilde H_{\Theta} = 0$, are actually solutions.
This was already taken into account in~\cite{Ober-Blobaum.Offen.2023.VariationalLearningEuler} and~\cite{Offen.Ober-Blobaum.2024.LearningDiscreteModels}, where authors learn Lagrangians from other variational integrators by minimizing this error.
To circumvent this, they add a penalization term on the Jacobian 
$$ J_\Theta(z_0, z_1, z_2; h) = D_{z_2} S_\Theta(z_0, z_1, z_2; h) , $$
encouraging either its determinant or its smallest singular value to be non-zero.
Our approach also combines an error term and a regularization term (weighed by some $\varepsilon > 0$), with the loss
\begin{equation}
\label{eq:LossDiscSch}
    \mathcal{L}_{\rm sch}(\Theta) = \frac{1}{|\mathcal{J}|} \sum_{j \in \mathcal{J}} \Bigl[ \bigl\| \bigl( J_\Theta^{(j)} \bigr)^{-1} S_\Theta^{(j)} \bigr\|^2_{\rm sch} + \varepsilon\, \log_{10} \bigl( \kappa (J_\Theta^{(j)}) \bigr) \Bigr] ,
\end{equation}
where $S_\Theta^{(j)}$ (resp. $J_\Theta^{(j)}$) is the one-step error (resp. its Jacobian) evaluated at the $j$-th data point $(z_0^{(j)}, z_1^{(j)}, z_2^{(j)})$.
The optimization is performed in batches.
Let us briefly explain this choice of loss function.

\bmhead{Error term}

Due to the multiscale nature of the problems we consider, we found it necessary to employ a similar strategy as in the continuous case~\eqref{eq:LossVFErr}.
By minimizing the one-step error directly, the multiscale components of the data are not captured, and no data-informed rescaling is possible.
We therefore minimize the error of Newton-Raphson iterations near the solution, which are homogeneous to the variations in phase-space and enable the use of a data-informed norm
\begin{equation}
\label{eq:normDiscSch}
    \| u \|_{\rm sch}^2 = u^\mathsf{T} M_{\rm sch}^{-1} u ,
\end{equation}
where the Gram matrix $M_\mathrm{sch} := \frac{1}{|\mathcal{J}|} \sum_{j \in \mathcal{J}} (z_2^{(j)} - z_1^{(j)}) (z_2^{(j)} - z_1^{(j)})^{\sf T}$ is the discrete equi\-valent of~\eqref{eq:normVFErr}. 
If this is zero, then so is the error of the scheme~\eqref{eq:oneStepError}, while avoiding the trivial quantities.

\bmhead{Regularization term}
This error term ensures the accuracy of the scheme, but may fail if the Jacobian stops being invertible.
To enforce this invertibility, we add a penalization term on its condition number, which is 
\begin{equation}
\label{eq:regulDiscrete}
    \kappa(J) = \left\| J \right\|\: \left\| J^{-1} \right\| .
\end{equation}
This also ensures that the linear system of Newton iterations are well-conditioned.
The logarithm ensures that this error term remains small even when the physical system presents large variations on the dataset.

This term is useful mostly at the start of the training, when the neural model varies a lot and the Jacobian may become degenerate. Unlike in VF learning where the regularization term needs to be minimized for the DVI to be well-behaved, here it seems to have no impact on the final learnt model and requires only a very small weight of $\varepsilon = 10^{-8}$.

\subsection{Using very large time-steps}
\label{subsec:timesteps}

The scheme~\eqref{eq:Scheme} being geometric means that formally, there exists a \textit{modified} one-form $z \mapsto \vartheta_h(z)$ and a \textit{modified} Hamiltonian $z \mapsto H_h(z)$.
Both are close to the original quantity up to $\mathcal{O}(h^2)$.

Owing to backward error analysis, we argue that the learnt dynamics are inaccurate \review{%
if they are not recovered using the same numerical scheme and time-step.
This is why the snapshots are assumed to be separated with uniform time-steps,\footnote{%
    \review{The time-step could be included as a parameter of the neural network to fit non-uniformly separated snapshots, but it is not the case here.}
} contrary to vector-field learning where adaptive time-steps can be used.
This does not mean that the scheme-fitted model is inaccurate in stiff region, as we shall see in~\S\ref{subsec:massless} with the massless charged particle test case.
}

To maximize efficiency, it seems fairly natural for long-time simulations to wish to choose $h$ very large. 
What could limit the size of the time-step with our approach?
\begin{enumerate}
    \item The modified quantities differ from the original ones up to an error of size $\mathcal{O}(h^2)$, and a large error may be compounded by iterations;
    \item The modified quantities $\vartheta_h$ and $H_h$ learnt by the network may only exist formally, meaning they can only be computed up to a flat function in $h\to 0$ (i.e. all the Taylor coefficients vanish at $0$); the approximation is then valid only for small $h$.
    \item During simulations, one must compute an initial guess of $z_{n+1}$. Standard approaches may yield guesses too inaccurate for optimisation techniques to solve the scheme, especially if it lands outside of the training set.
\end{enumerate}
For the first remark, the main risk is to learn an inaccurate symplectic form, which would yield a non-symplectic scheme---this is different from SympNets, which work with a known symplectic form\review{, but is similar to Poisson Neural Networks (PNNs) \citep{Jin.Zhang.ea.2020.LearningPoissonSystems} for which the symplectic form is unknown.}
Also, much like SympNets and PNNs, there is indeed no guarantee that the Hamiltonian is exactly preserved along iterations: there might be a drift due to compounding errors. 
This drift exists for every method of learning, and should not invalidate our approach. 
For the second remark, it is important to known that these formal results are an ``at worst'' case, and it may still be beneficial to evaluate this approach for specific test cases. 
For example, let us ignore structure and consider the Euler scheme applied to $\dot y = f(y)$ of flow $(t,y_0) \mapsto \varphi^t(y_0)$. 
The modified vector field is $f_h := \frac{1}{h}(\varphi^h - \mathrm{id})$, which is well-defined for all timesteps.
Finally, for the third remark, it may be useful to use a neural network (NN) for the initial guess.
This baseline NN should not require a particular structure, but may be invertible~\citep{Ardizzone.Kruse.ea.2019.AnalyzingInverseProblems}.

\section{Numerical experiments}
\label{sec:simulations}

The different learning strategies are tested on the same Neural Degenerate Lagrangian architecture: the symplectic one-form and the Hamiltonian are different MLPs, each with their own parameters, and normalized inputs.
The optimizer is always Adam.

\subsection{Lotka-Volterra}
\label{subsec:LotkaVolterra}

We consider here the same Lotka-Volterra problem as the one introduced in~\S\ref{subsec:long_time_intro}, where the difficulties of the learning strategies have been exposed. 
We therefore choose not to recall them here, but this time we will follow the approaches described in Sections~\ref{sec:learnVF} and~\ref{sec:learnNum}.

\bmhead{Hyperparameters}
The dataset consists in 2000 trajectories, separated between a training, a test and a validation set, representing 80\%, 15\% and 5\% of the data respectively. 
As in~\S\ref{subsec:long_time_intro}, their initial conditions are uniformly sampled in the subset of phase space such that $H(x, y) \leq 4.4$, and evolve over 5 time-steps with $h = 0.1$. 
This time-step and maximum energy are chosen such that trajectories at the edge of the domain are barely stable when using the DVI scheme.

Each network $\vartheta_\Theta$ and $H_\Theta$, independently of the learning strategy, consists of three layers of size 30, and the inputs are normalized as described in~~\S\ref{subsec:nns}. 
The models undergo four trainings, first during 20 epochs with a learning rate of $10^{-2}$, then 500 epochs with $\mathrm{lr}$ $10^{-3}$ and finally two trainings of 500 epochs with $\mathrm{lr}$ $10^{-4}$.
The regularization parameter is set to $\varepsilon = 10^{-6}$ for VF learning (\S\ref{sec:learnVF}) and $\varepsilon = 10^{-8}$ for scheme learning (\S\ref{sec:learnNum}).

\begin{figure}
    \centering
    \includegraphics[width=.99\textwidth]{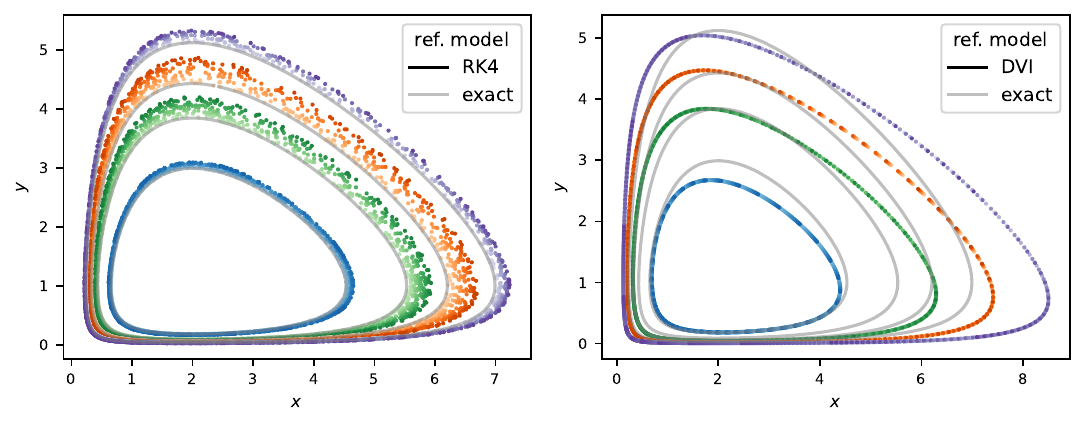}%
    \vspace{-.2cm}%
    \caption{(Lotka-Volterra) Numerical solutions of the reference model \eqref{eq:LotkaVolterraDiffEq} with different initial data. Left: RK4 solver with time-step $h = 0.1$ and 250k steps. Right: DVI solver with time-step $h = 0.1$ and 100k steps. Points are displayed every 51 steps. Exact solutions are refined numerical solutions of the reference model.
    \review{The legend only denotes the line or marker style, different colors indicate different trajectories.}
    }
    \label{fig:lv:longtimeRef}
\end{figure}

\begin{figure}
    \centering
    \includegraphics[width=.99\textwidth]{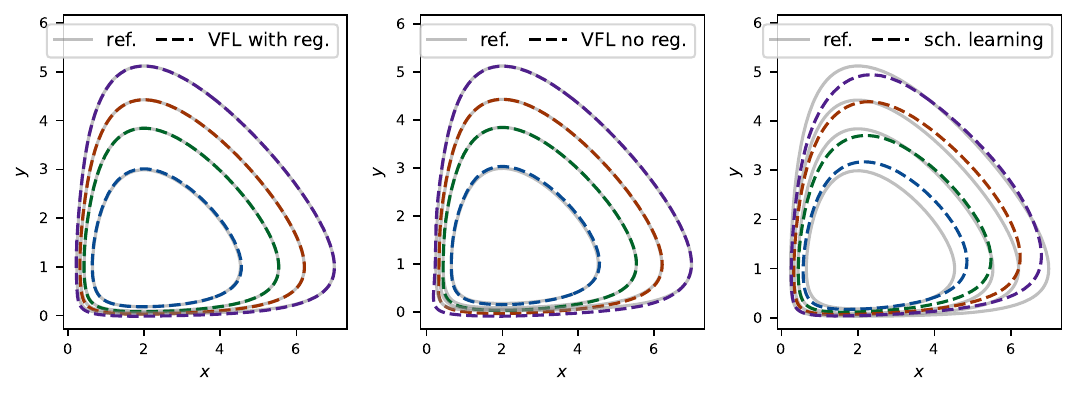}%
    \vspace{-.2cm}%
    \caption{(Lotka-Volterra) \review{\emph{The exact flow of scheme learning is inaccurate.}} Refined numerical solutions for each neural model (dashed lines), compared with the refined solutions of the reference model (solid lines), obtained by using \texttt{solve\_ivp}. In reading order: VF learning (VFL) with and without regularization, and scheme learning.
    \review{The legend only denotes the line style, different colors indicate different trajectories.}
    }
    \label{fig:lv:nn_ex}
\end{figure}

\begin{figure}
    \centering
    \includegraphics[width=.499\textwidth]{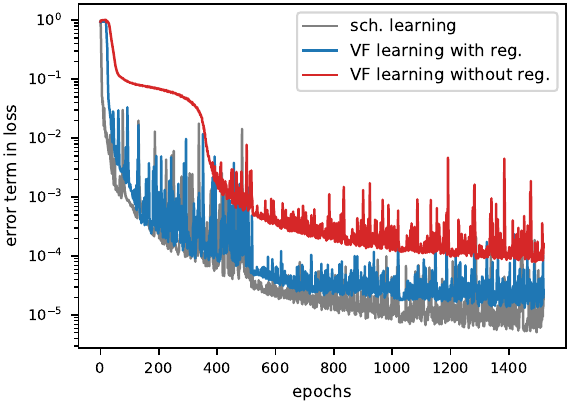}\hfill%
    \includegraphics[width=.499\textwidth]{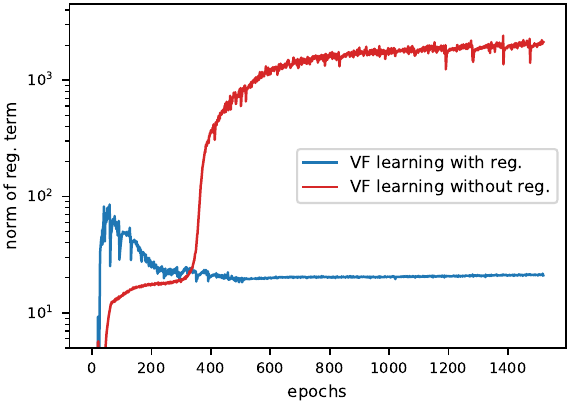}%
    \caption{(Lotka-Volterra) \emph{\review{Blow-up of the regularisation term without penalization.}} Evolution of the loss and the regularization term on the test dataset throughout the training.
    }
    \label{fig:lv:lossTrace}
\end{figure}

\bmhead{Results of vector-field learning}
In Figure~\ref{fig:lv:nn_ex}, we compute highly-accurate solutions of the neural models using the \texttt{solve\_ivp} routine of Scipy with parameters \texttt{rtol=1e-10} and \texttt{atol=1e-12} over a single period, and compare them to exact solutions of the reference model that are computed the same way.
Over short times, both VF-learning neural networks closely match the reference model, though the regularized model is slightly more accurate near the origin.
This matches the training metrics in Figure~\ref{fig:lv:lossTrace}, where the error term in the loss is slightly larger for the non-regularized model, but both are small (of order $10^{-4}$).
It appears that our new regularization term has no negative impact on the accuracy of the neural model.

\begin{figure}
    \centering
    \includegraphics[width=.99\textwidth]{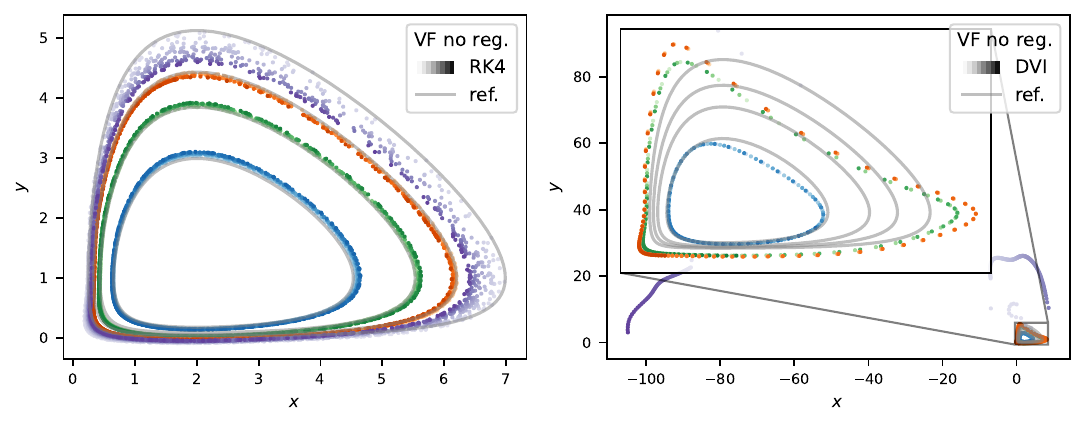}%
    \vspace{-.25cm}%
    \caption{(Lotka-Volterra) \review{\emph{Instability of the DVI without regularisation.}} Numerical solutions  of the non-canonical neural model, obtained without regularization, for different initial conditions. Left: RK4 solver with time-step $h = 0.1$ for 250k steps. Right: DVI solver with time-step $h = 0.1$ for 200 steps. Points are displayed every 51 steps for the RK4 solution, every step is shown for the DVI solution. Reference solutions are refined numerical solutions of the reference model.}
    \label{fig:lv:longtimeCMY}
\end{figure}

\begin{figure}
    \centering
    \includegraphics[width=.99\textwidth]{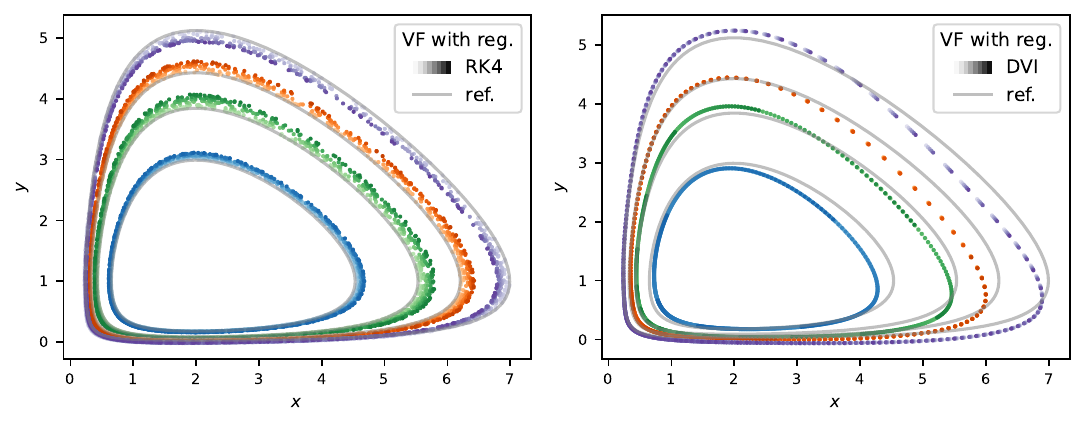}%
    \vspace{-.25cm}%
    \caption{(Lotka-Volterra) \review{\emph{Effect of the regularisation on the DVI solution.}} Numerical solutions of the VF-learning neural model, obtained with regularization, for different initial conditions. Left: RK4 solver with time-step $h = 0.1$ for 250k steps. Right: DVI solver with time-step $h = 0.1$ for 100k steps. Points are displayed every 51 steps for the RK4 solution, every step is shown for the DVI solution. Reference solutions are refined numerical solutions of the reference model.}
    \label{fig:lv:longtimeVF}
\end{figure}

When observing the long-time behavior of both models in Figures~\ref{fig:lv:longtimeCMY} and~\ref{fig:lv:longtimeVF} (and the reference model in Figure~\ref{fig:lv:longtimeRef}), 
the qualitative properties of the numerical schemes still hold.
Over short time intervals, the DVI is less accurate than RK4 
due to its first-order nature and the large time step used. However, over longer time periods, 
the DVI maintains a periodic solution that does not degrade, 
unlike the RK4 scheme which produces a dissipative solution. 
Because of this periodicity, we stop after 100k time-steps for the (more costly) DVI, while we perform 250k time-steps for the RK4 scheme.

Comparing the behavior of the DVI on both VF-learning models, 
it is clear that the penalization method successfully addresses 
the gauge issue identified in Section 2 (Figure \ref{fig:perturbedLotkaVolterra} on the right): without it, the DVI is unstable, and with it, the DVI is more accurate than on the reference model.
This matches the trace of the regularization term $r_\Theta$ during training in Figure~\ref{fig:lv:lossTrace} (with weight $\varepsilon = 0$ for the model ``without regularization''), which is much larger if it is not included in the loss.
This further supports the idea that learning the structure alone is not sufficient. One must 
also use a time integration scheme that is adapted to this structure and, in this case, take this scheme into account during the training process.

\begin{figure}
    \centering
    \includegraphics[width=.99\textwidth]{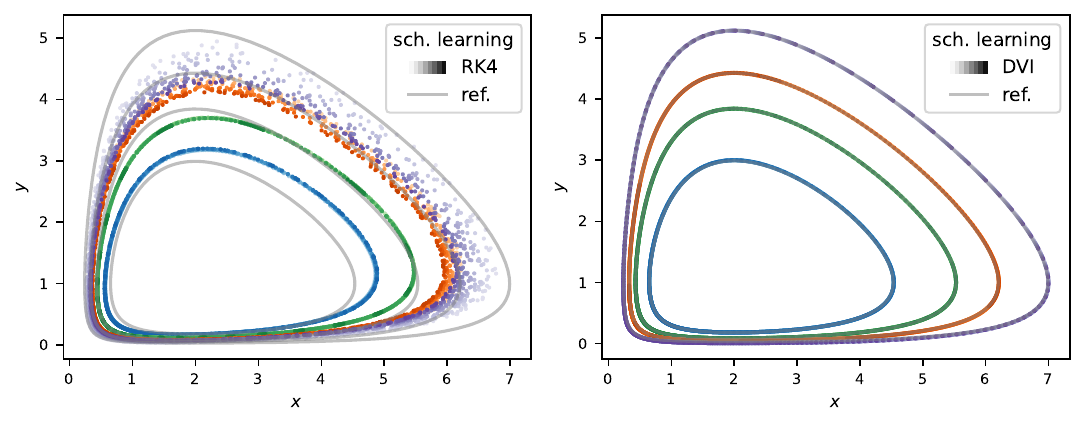}%
    \vspace{-.25cm}%
    \caption{(Lotka-Volterra) \review{\emph{Accuracy of the DVI after scheme learning.}} Numerical solutions  of the scheme-learning neural model for different initial conditions. Left: RK4 solver. Right: DVI solver. They are obtained using a time-step $h = 0.1$ (same as in the dataset) for 100k steps. Points are displayed every 31 steps. Reference solutions are refined numerical solutions of the reference model.}
    \label{fig:lv:longtimeSch}
\end{figure}

\begin{figure}
    \centering
    \includegraphics[width=.99\textwidth]{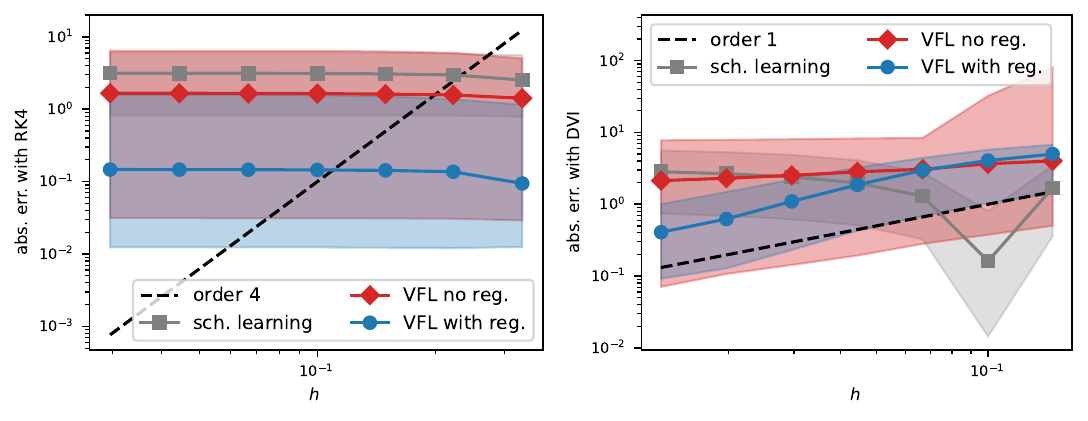}%
    \vspace{-.25cm}%
    \caption{ (Lotka-Volterra) Short-time numerical error for the different learning strategies with the RK4 scheme (left) and the first-order DVI (right), as measured on the validation set up to $t = 10.0$. The values drawn being the median (solid line), as well as the 5th and 95th percentiles (filled area).
    }
    \label{fig:lv:numErr}
\end{figure}

\bmhead{Results of scheme learning}
The simulations in Figure~\ref{fig:lv:longtimeSch} using the DVI demonstrate that the scheme-learning corrects the error of the DVI, generating highly accurate closed solutions over long times.
This suggests that the training process learns a modified potential and Hamiltonian that compensate for the scheme-induced error. 
Since the DVI scheme is used during training to perform time stepping, 
the results are consistent with those reported in \cite{Jin.Zhang.ea.2020.SympNetsIntrinsicStructurepreserving}, where learning a discrete flow was shown to yield higher accuracy than learning the underlying ODE. 

However, it is important to note that the learned model is specific to the scheme and time step used during training, 
and its performance is likely to degrade when either is altered.
The short-time error analysis in Figure~\ref{fig:lv:numErr} shows that the median numerical error is lowest when using the DVI with time-step $h = 0.1$, which matches the training data. 
When using smaller time-steps, the error degrades and plateaus while the VF-learning models converge with order~1 as expected (up to a learning error).
The numerical solution using RK4 scheme is also much less accurate, even with a time-step $h = 0.1$.
This can also be seen in Figure~\ref{fig:lv:nn_ex}, where the exact solution of the scheme-learning model is much less accurate than the numerical solution yielded by the DVI in Figure~\ref{fig:lv:longtimeSch}.

\subsection{Massless charged particle}
\label{subsec:massless}

Consider a standard Lorentz force in two dimensions ($z$ the 2D space variable and $v$ the 2D velocity) with magnetic potential $z \mapsto A(z)$ and electric potential $z \mapsto \varphi(z)$, applied to a particle of mass $m$ and charge $q$. 
Classically, the Lagrangian is $(z, v) \mapsto \frac12 m |v|^2 + q A(z)\cdot v - q\varphi(z)$. 
Writing $z = (x, y)$ \review{and $A = (A_x, A_y)$}, the magnetic potential can be chosen as $(\vartheta, 0)$ and still generate the same magnetic field $B = \partial_x A_y - \partial_y A_x = -\partial_y \vartheta$, setting $\vartheta(x, y) = A_x(x, y) - \int_0^y \partial_x A_y(x, \mathrm{y}) \mathrm{d}\mathrm{y}$.
In the limit $m \to 0$, the kinetic energy disappears, the Lagrangian can be rescaled $L \leftarrow L / q$ and the velocity can be replaced by $z_t$, 
\begin{equation*}
    L(x, y, x_t) = \vartheta(x, y) x_t - \varphi(x, y) .
\end{equation*}

For this experiment, the magnetic potential is chosen to increase from the origin as $A(x, y) = \frac{A_0}{2} (1 + x^2 + y^2) (-y, x)^{\sf T}$, i.e.
\begin{equation}
    \vartheta(x,y) = -A_0\, y \left( 1 + 2x^2 + \tfrac{2}{3} y^2 \right),
\end{equation}
which generates a magnetic field $B(x, y) = A_0 (1 + 2x^2 + 2y^2)$.
The electrostatic potential is
\begin{equation}
    \varphi(x, y) = E_0 \bigl(2 - \cos(x) - \sin(y) \bigr) .
\end{equation}
According to \eqref{eq:degenerate_vf}, this yields the dynamics $\dot x = - \partial_y \varphi / B$ and $\dot y = \partial_x \varphi / B$, i.e.
\begin{equation}
\label{eq:model_Massless}
    \dot{x} = \frac{E_0 \cos(y)}{A_0(1 + 2x^2 + 2y^2)}, \qquad
    \dot{y} = \frac{E_0 \sin(x)}{A_0(1 + 2x^2 + 2y^2)}.
\end{equation}
This is coherent with the classical Lorentz dynamics $m \dot v = q\bigl( E(z) + \dot z \times B(z) \bigr)$ in the limit $m \to 0$, with $E = -\nabla\varphi$ and $B = \nabla\times A$.
In the sequel, we set $A_0 = 1$ and $E_0 = 1$.

\bmhead{Hyperparameters}
We generate 15k initial conditions $(x_0, y_0)$ in the ball of radius $\pi$ centered at $(0, \pi/2)$ such that the electrical energy is not greater than 1.5, i.e. in the set $\{(x, y) \in B_\pi(0, \pi/2) \mid \varphi(x, y) \leq 1.5\}$.
This is done using a Latin hypercube method in polar coordinates around $x = 0, y = \pi/2$ and filtering the data based on the value of $\varphi$.
For scheme-learning, the time-step is set to be $h = 0.5$.

Each network $\vartheta_\Theta$ and $H_\Theta$, independently of the learning strategy, consists of two layers of size 50, and the inputs are normalized. 
The models undergo three trainings, first during 20 epochs with a learning rate of $10^{-2}$, then 500 epochs with $\mathrm{lr}$ $10^{-3}$ and finally 500 epochs with $\mathrm{lr}$ $10^{-4}$.
The regularization parameter is set to $\varepsilon = 10^{-4}$ for VF learning and $\varepsilon = 10^{-8}$ for scheme learning. 

\begin{figure}
    \centering
    \includegraphics[width=0.49\linewidth]{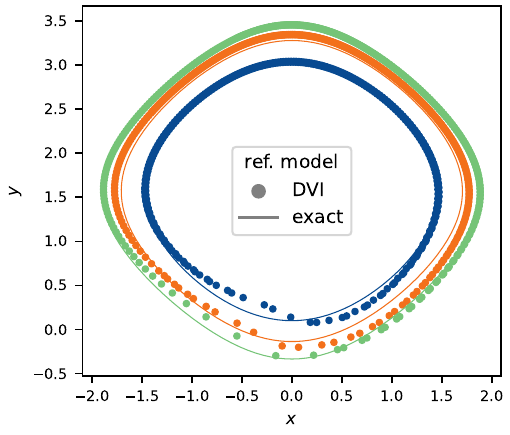}\hfill%
    \includegraphics[width=0.49\linewidth]{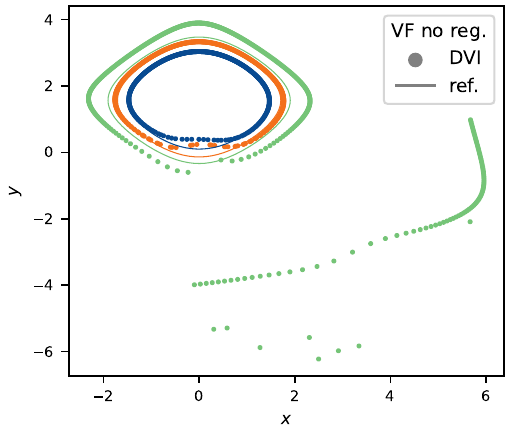}
    \includegraphics[width=0.49\linewidth]{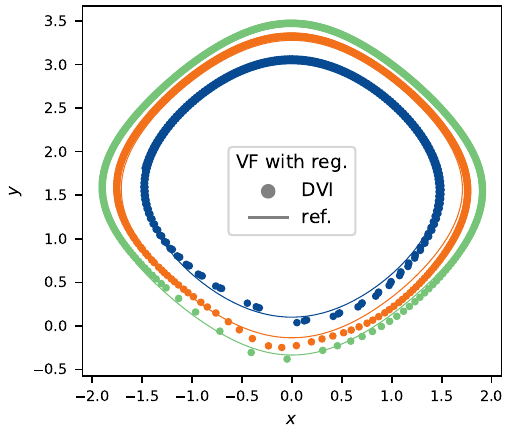}\hfill%
    \includegraphics[width=0.49\linewidth]{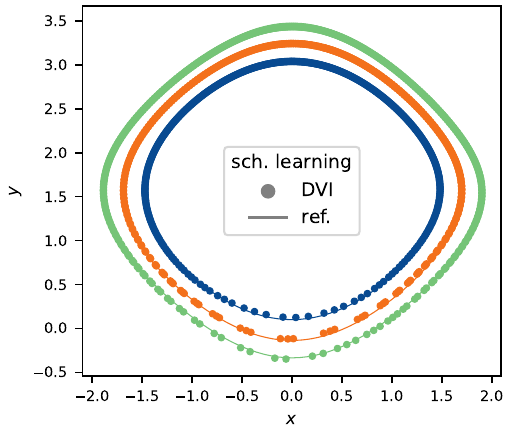}
    \caption{(Massless charged particle) \emph{\review{Confirmation of previous results.}} Simulations of the reference model \eqref{eq:model_Massless} (top left) VF-learning model trained without regularization (top right), VF-learning model with regularization (bottom left), scheme-learning model (bottom right),  with the DVI scheme on three different trajectories with time-step $h = 0.5$ for 500 time-steps.
    \review{The legend only denotes the line or marker style, different colors indicate different trajectories.}
    }
    \label{fig:masslessDVI}
\end{figure}

\bmhead{Results}
In the top left part of Figure \ref{fig:masslessDVI}, we can observe that the reference model \eqref{eq:model_Massless}, solved with the DVI scheme, 
exhibits a periodic behavior for the three trajectories.
Since the time step is chosen very large, we observe some phase error. 
On the top right part, we see that the VF-learning model on one trajectory (the most external one) generates a completely wrong solution and the periodicity is lost. As with the Lotka-Volterra problem, it is very likely that a modified potential have been learned.

\begin{figure}
    \centering
    \small{VF with reg.\qquad\qquad\qquad  VF without reg.\qquad\qquad\qquad sch. learning}
    \includegraphics[width=.98\textwidth]{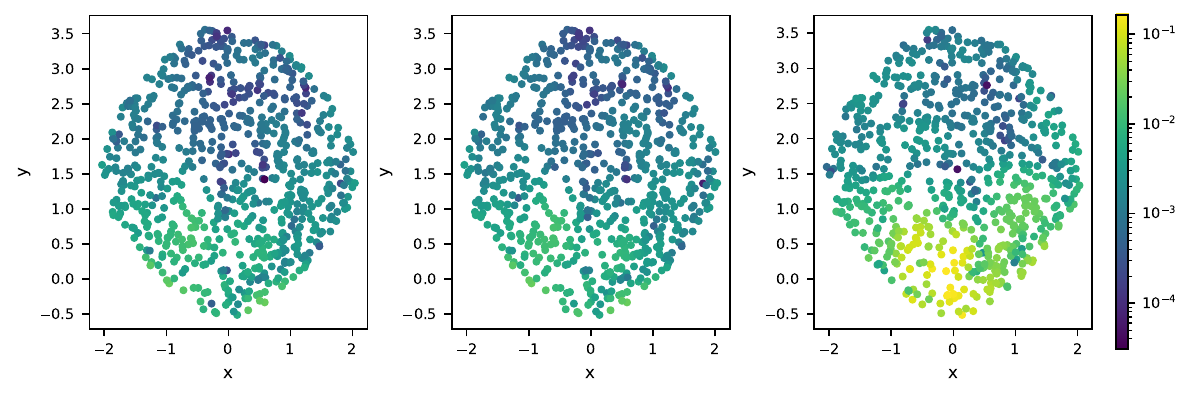}
    \caption{(Massless charged particle) Loss values on the validation set for the VF-learning models with (left) or without (center) regularization, and for the scheme-learning model (right).}
    \label{fig:masslessErrors}
\end{figure}

On Figure \ref{fig:masslessDVI} (bottom left), we can observe that the regularization allows us, as before, to capture the correct trajectories, even though the learning error (in phase-space, Figure~\ref{fig:masslessErrors}) is very similar to the non-regularized model.
Although not identical to the reference solutions, the errors appear to be of the same order as when DVI 
is applied to the real system with a large time step. The dominant error therefore seems to come from the scheme rather than the learning.

On the bottom right of Figure~\ref{fig:masslessDVI}, we can see that the approach that uses the scheme to learn, thus capturing an efficient discrete flow, partially corrects the DVI error. 
Even though the learning error (Figure~\ref{fig:masslessErrors}) is highest where the DVI seems to be most inaccurate (near the origin), the numerical solutions are closer to the reference solutions when using this model.
This result is comparable to those obtained for the Lotka-Volterra system, which validates the previous conclusions.

\subsection{Guiding center}

This test case is an asymptotic model of plasma physics in tokamaks with a strong magnetic field. 
The position is expressed in poloidal-toroidal coordinates $X = (r, \theta, \phi)$, where $r$ is the minor-radial position, $\theta$ the geometric poloidal angle, and $\phi$ the geometric toroidal angle. 
The momentum is reduced to a single coordinate $u$ in the toroidal direction, parallel to the magnetic field. 
Owing to the particular geometry and reduced dynamics, the problem is non-canonical and its Lagrangian is commonly written 
\begin{equation*}
    L(X,u, X_t) = \bigl( A(X) + u b(X) \bigr)\cdot X_t - H(X,u) ,
\end{equation*}
where $A = (0, A_\theta, A_\phi)$ is a magnetic potential from which the vector field ${\bf B} = \nabla\times A$ is derived, and $b = {\bf B} / \|{\bf B}\|$ is the magnetic field unit vector.

We assume that the magnetic field is only in the toroidal direction, i.e. $ub(X) \cdot X_t = u \bigl( R_0 + r\cos(\theta) \bigr) \phi_t$. 
Reordering the coordinates into $z = (\theta,\phi,r,u)$, the Lagrangian is then properly degenerate, i.e. it admits the $(x, y)$ decomposition.\footnote{
    If the problem is not properly degenerate, then it is possible to define an equivalent Lagrangian which is, as in~\cite{Qin.Guan.ea.2009.VariationalSymplecticAlgorithm}. 
    This is done by recognizing an identity $A_r(X) \dot r = \frac{{\rm d}}{{\rm d}t} f(X) - (\partial_\theta f) \dot \theta - (\partial_\phi f) \dot\phi$ for a well-chosen $f$, and removing $\frac{\mathrm{d}}{\mathrm{d}t} f(X)$ by invariance.
}
Following~\cite{Ellison.Finn.ea.2018.DegenerateVariationalIntegrators}, we write
\begin{equation*}
    L(\theta,\phi,r,u, \theta_t, \phi_t) = A_\theta(r,\theta) \theta_t + \bigl(A_\phi(r) + u (R_0 + r \cos\theta) \bigr) \phi_t - H(r,\theta,u) .
\end{equation*} 
with the symplectic (or magnetic) potential 
\begin{equation}
    \label{eq:magneticPotential}
    A_\theta(r,\theta) = \frac{B_0 R_0^2}{\cos^2(\theta)} \left( \frac{r \cos(\theta)}{R_0} - \log\left(1 + \frac{r \cos(\theta)}{R_0}\right) \right), \qquad
    A_\phi(r) = -\frac{B_0 r^2}{2q_0}.
\end{equation}
The parameters are $R_0$ the radial position of the magnetic axis; $B_0$ the magnitude of the magnetic field at $R_0$; $q_0$ the (dimensionless) safety factor, regarded as constant.
The Hamiltonian is a combination of kinetic and magnetic energies,
\begin{equation}
\label{eq:gc:Hamiltonian}
    H(r,\theta,u) = \frac{1}{2}u^2 + \mu B(r,\theta), \qquad\quad
    B(r,\theta) = \frac{B_0}{1 + \frac{r \cos(\theta)}{R_0}} \sqrt{1 + \left( \frac{r}{q_0 R_0} \right)^2} ,
\end{equation}
where the additional parameter $\mu$ is the (constant) magnetic moment of the particle and $B = \|{\bf B}\|$ is the magnitude of the magnetic field. 
In the sequel we set $B_0 = 1$, $R_0 = 1$, $\mu = 2.25\cdot 10^{-6}$ and $q_0 = 2$.

\begin{figure}
    \centering
    \includegraphics[width=.99\textwidth]{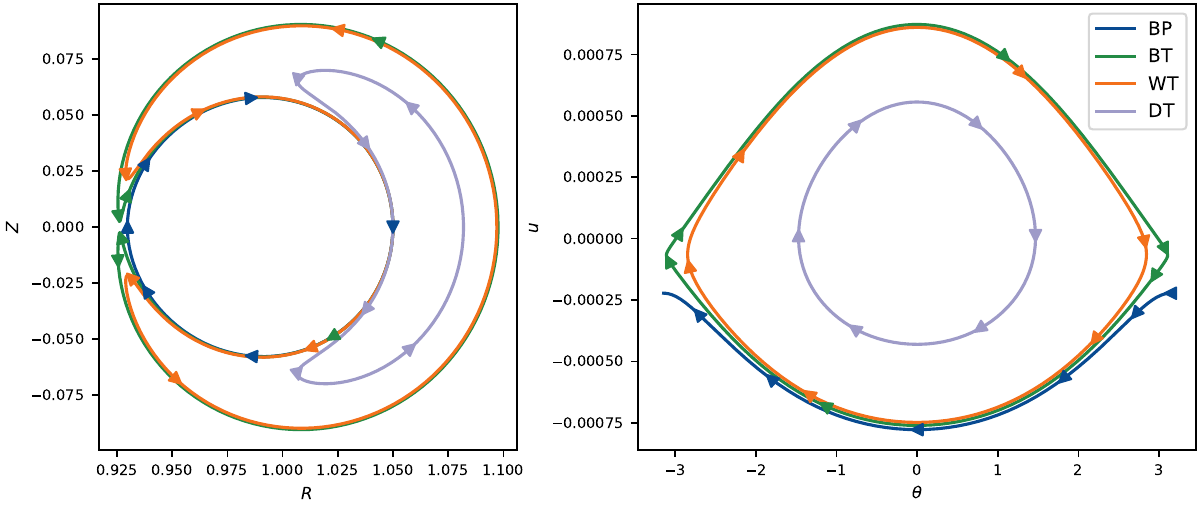}%
    \caption{(Guiding center) The trajectories of interest in the poloidal $(R, Z)$  plane (left) and in the $(\theta, u)$ plane (right). The initial conditions are $r_0 = 0.05$, $\theta_0 = 0$, $\phi_0 = 0$ and varying $u_0$, from smallest to largest in the legend (or, in absolute value, from largest to smallest).}
    \label{fig:trajGuidingCenter}
\end{figure}

\bmhead{Reference behaviour}
We are interested in four trajectories, with the same initial position $r_0 = 0.05$, $\theta_0 = 0$ and $\phi_0 = 0$, and different initial velocities:
\begin{itemize}
    \item \textbf{BP} -- barely passing trajectory, $u_0 = -7.782 \cdot 10^{-4}$;
    \item \textbf{BT} -- barely trapped trajectory, $u_0 = -7.610 \cdot 10^{-4}$;
    \item \textbf{WT} -- well trapped trajectory, $u_0 = -7.487 \cdot 10^{-4}$;
    \item \textbf{DT} -- deeply trapped trajectory, $u_0 = -4.306 \cdot 10^{-4}$ (also called ``banana orbit'').
\end{itemize}
As can be seen in Figure~\ref{fig:trajGuidingCenter} with the BP trajectory, if the initial velocity is large enough in absolute value, then, similarly to a pendulum, the velocity does not change sign, i.e. the particle never ``turns back'' in the tokamak (hence the terminology of \textit{passing} vs \textit{trapped}).
In the poloidal plane, also called the $(R, Z)$ plane with $R = R_0 + r \cos(\theta)$ and $Z = r \sin(\theta)$, the trajectory changes from a crescent shape to a nearly circular orbit.
Since the plots in $(R, Z)$ and in $(\theta, u)$ are closely related, we will only plot the former when evaluating our neural networks.

\begin{figure}
    \centering
    \includegraphics[width=.99\textwidth]{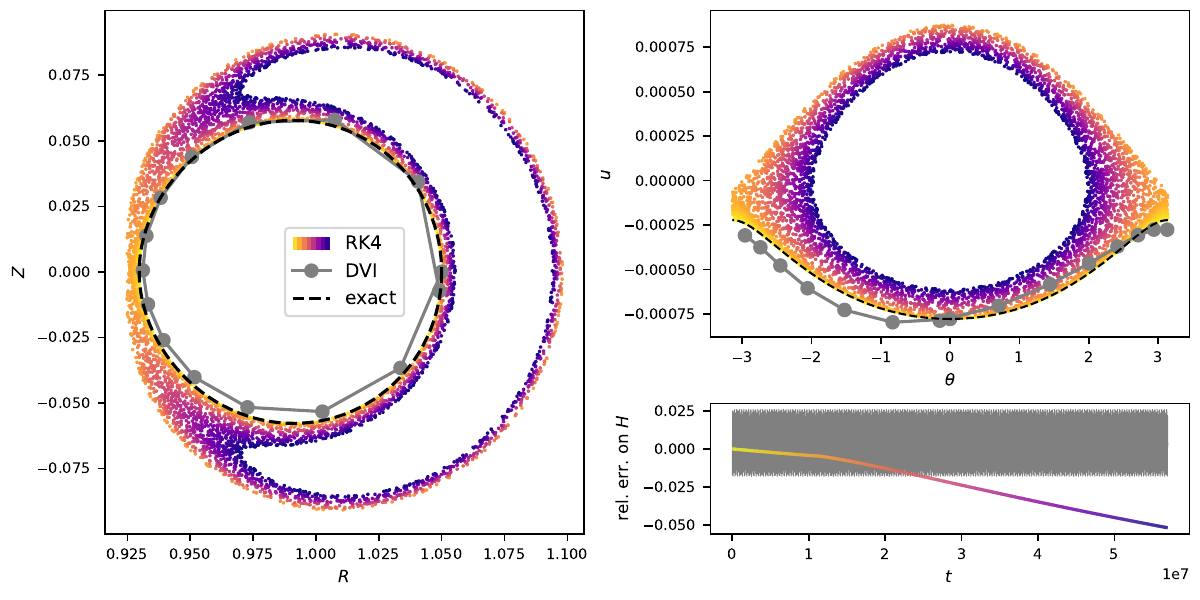}%
    \vspace*{-.1cm}%
    \caption{(Guiding center) \emph{\review{Energy dissipation of RK4.}} Behaviour of the numerical solutions applying the RK4 or the DVI schemes to the reference model, with time-step $h = T_\mathrm{DT} / 20$ up to time $t = 1500 T_\mathrm{DT}$. 
    In phase space, the RK4 solution is plotted every 5 time-step, and only the first period is plotted for the DVI scheme.}
    \label{fig:refRK4GuidingCenter}
\end{figure}

\begin{figure}
    \centering
    \includegraphics[width=.49\textwidth]{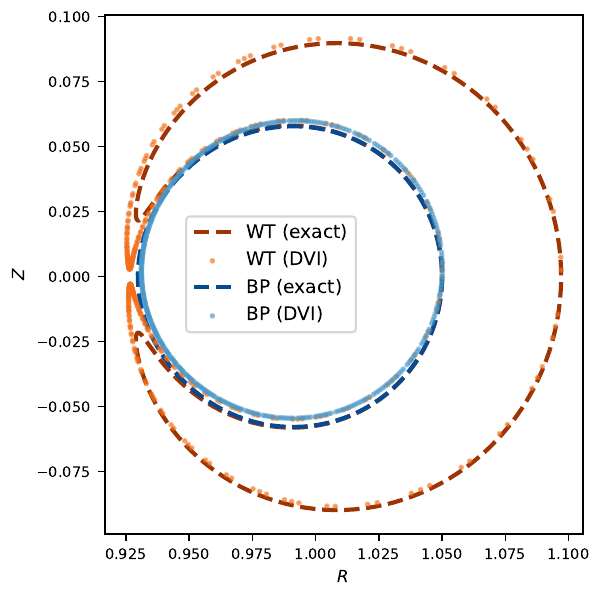}\hfill%
    \includegraphics[width=.49\textwidth]{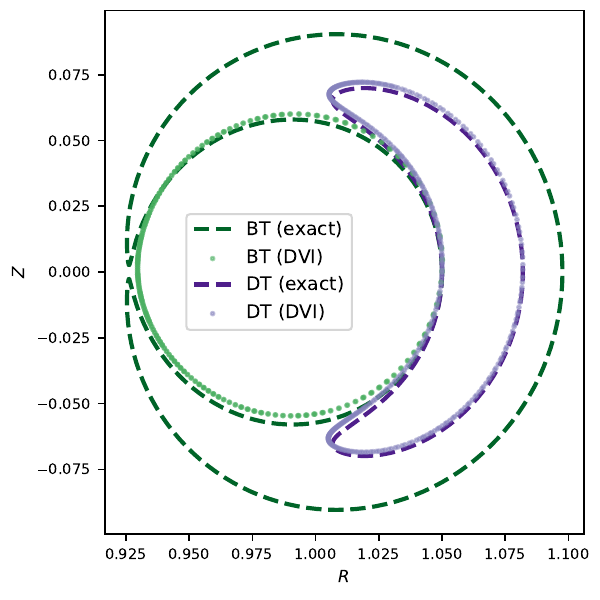}%
    \vspace*{-.1cm}%
    \caption{(Guiding center) \emph{\review{DVI solutions on validation trajectories.}} Comparison on the reference model between the orbits from of the exact solution and of the numerical solution obtained with the DVI using a time-step $h = T_\mathrm{DT} / 20$.
    }
    \label{fig:charactRefGuidingCenter}
\end{figure}

To perform simulations, we choose 20 time-steps per period of the DT trajectory, i.e. $h = T_\mathrm{DT} / 20$ with $T_\mathrm{DT} = 37974.6$. 
Over long-times, using the RK4 scheme fails to capture the passing behaviour (Figure \ref{fig:refRK4GuidingCenter}). 
Using the DVI scheme (Figure \ref{fig:charactRefGuidingCenter}), the BT trajectory is be wrongly categorized as ``passing.'' 
This test case shows that even without considering very long time spans, it is crucial to use a structure-preserving scheme to capture the full set of physical solutions.
Learning while preserving the structure may not be sufficient in practice.

\bmhead{Pre-processing of the input}

We hard-code axisymmetry, the independence of the output w.r.t. $\phi$, by discarding this coordinate from the inputs. 
Additionally, to take into account the angular nature of $\theta$, we apply a transformation from~\cite{Dong.Ni.2021.MethodRepresentingPeriodic}. 
Each coordinate is repeated $k = 6$ times, $v \in \mathbb{R} \mapsto v \mathbbm{1} := (v, ..., v)^{\sf T} \in \mathbb{R}^k$, the angles are shifted and only their cosine is kept.
This yields a pre-processing layer of the form
\begin{equation*}
    \begin{pmatrix}
        \theta \\ \phi \\ r \\ u
    \end{pmatrix} \mapsto 
    \begin{bmatrix}
        \cos_{\odot}(\theta \mathbbm{1} + \varphi_\Theta) \\
        r\mathbbm{1} \\
        u\mathbbm{1}
    \end{bmatrix} ,
\end{equation*}
where $\varphi_\Theta \in \mathbb{R}^k$ are parameters to be learnt (initialised randomly, uniformly in $[0, 2\pi]$) and $\cos_{\odot}$ is the cosine applied componentwise.
The inputs are then normalized such that all coordinates are in $[0, 1]$, as for other problems.

The output of the Hamiltonian is also rescaled with an affine function which maps $[0, 1]$ to $\left[ \min \|\dot z\|, \max \|\dot z\| \right]$, the extrema being determined on the training set. 
For scheme learning (\S\ref{sec:learnNum}), this is replaced by the finite-difference $\frac{1}{h} \| z_{n+1} - z_n \|$.

\bmhead{Training details}
We uniformly generate 600 initial conditions in the tokamak with $r_0^{\ 2} \in [0.03^2, 0.055^2]$, $\theta_0 \in \left[-\frac{\pi}{10}, \frac{\pi}{10} \right]$, $\phi_0 \in [0, 2\pi]$ and $u_0 \in \left[ -9\cdot 10^{-4}, -3\cdot 10^{-4} \right]$. 
Each trajectory then evolves for 60 time-steps with $h = T_\mathrm{DT}/20$, and is assigned to either the training, test or validation set with ratios of 80\%, 15\% and 5\% respectively. 
This construction provides some ``padding'' around the specific trajectories we are interested in.

Each network consists of 3 layers of dimension 48, meaning there are 5.8k parameters in $\vartheta_\Theta$ and 5.8k parameters in $H_\Theta$, for a total of about 12k parameters per model.\footnote{
    Because there are 2 outputs for $\vartheta_\Theta$ and only 1 for $H_\Theta$, there are 48 more parameters in $\vartheta_\Theta$ (there is no final bias for either network).
}
The models are trained over 3 training: 20 epochs with learning rate $10^{-2}$, 500 with $\mathrm{lr}$ $10^{-3}$ and 500 with $\mathrm{lr}$ $10^{-4}$.
The regularization weight is $\varepsilon = 1$ for VF learning (though we will observe no difference when setting $\varepsilon = 0$), and $\varepsilon = 10^{-8}$ for scheme learning. 

Additionally, the first two trainings use an approximation of~\eqref{eq:oneStepError} in order to reduce the number of function evaluations.
This uses a first-order Taylor expansion of $\vartheta(x_2, y_2)$ around $(x_1, y_1)$ and is written
\begin{multline}
\label{eq:modifOneStepError}
    \widetilde{J}_h^{-1} \widetilde{S}_h(z_0,z_1,z_2) =\\ 
    \begin{bmatrix} 
        y_2 - y_1 + (D_y\tilde{\vartheta}_\Theta)^{-1} \left(D_x\tilde\vartheta_\Theta (x_2 - x_1) - (D_x\tilde \vartheta_{\Theta})^\mathsf{T} (x_1 - x_0) + h \nabla_x \tilde H_{\Theta}\right) \\
        x_1 - x_0 - h (D_y\tilde{\vartheta}_\Theta)^{-\mathsf{T}} \nabla_y \tilde H_{\Theta}
    \end{bmatrix} ,
\end{multline}
where every $\tilde \vartheta_\Theta$ and $\tilde H_\Theta$ are evaluated at $(x_1, y_1)$. 
Because this requires only 1 function evaluation instead of 2, and uses a smaller matrix inversion (only on $D_y\vartheta_\Theta$), this greatly speeds up the otherwise slow training.

\begin{figure}
    \centering
    \includegraphics[width=.99\textwidth]{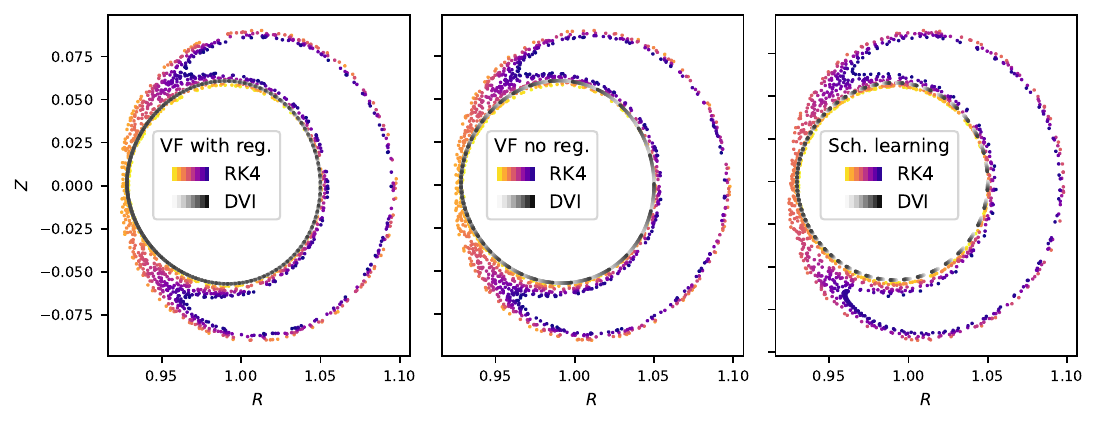}%
    \vspace{-.1cm}%
    \caption{(Guiding center) Numerical solutions for the BP trajectory of the different neural models.}
    \label{fig:gc:bp_nn}
\end{figure}

\begin{figure}
    \centering
    \includegraphics[width=.99\textwidth]{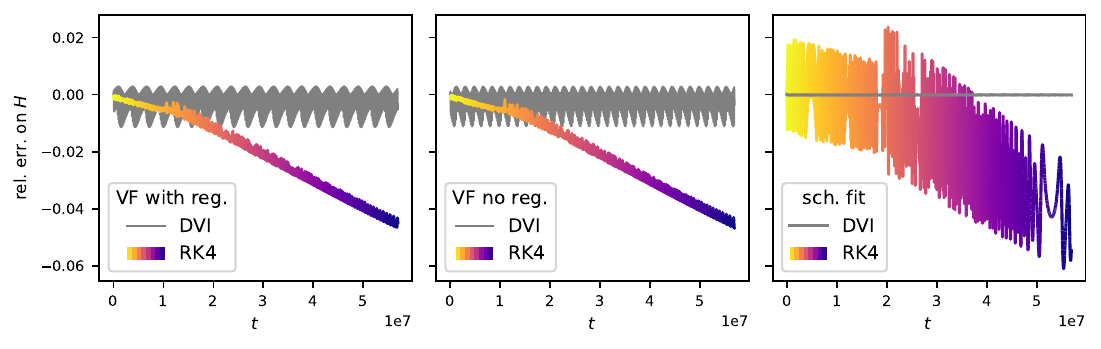}%
    \vspace{-.1cm}%
    \caption{(Guiding center) Evolution of the relative error on $H$ for the BP trajectory of the different neural models.}
    \label{fig:gc:err_h}
\end{figure}

\bmhead{Behaviour of the models}
In Figure~\ref{fig:gc:bp_nn}, it appears that the numerical solutions of the barely passing (BP) trajectory behave qualitatively the same as for the reference model.
Using the RK4 scheme, the solution loses energy over time, ending up in a trapped trajectory.
This is confirmed in Figure~\ref{fig:gc:err_h}, where this change of regime is seen in the slope of the energy decrease.
As for the other test cases, the scheme-learning model is less accurate than the VF-learning models with this numerical scheme: the energy oscillates more.
However, it oscillates less when using the DVI, a difference of $10^2$ in amplitude.

The center plots show that for this particular test case, the issue of gauge invariance does not appear. 
The solution of the DVI scheme behaves well even for the model trained without the regularization term.
In fact, when monitoring the regularization term during training, both VF-learning models have similar values. 
This may be due to the hard-coding of the axisymmetry, to the relatively small size of the networks or to the fairly short training, which causes the maps to be smoother than with simpler problems.
We now focus on the model trained \textit{with} the regularization term to verify that even when including this term, the learnt dynamics are accurate.


\begin{figure}
    \centering
    \includegraphics[width=0.999\textwidth]{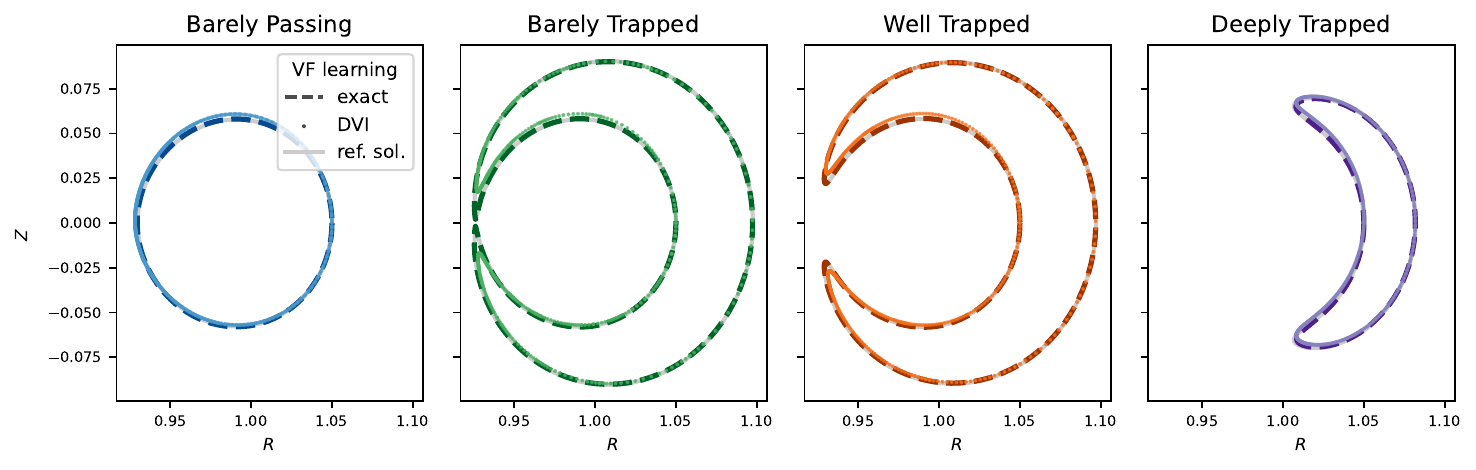}%
    \caption{(Guiding center - VF-learning) Comparisons of the exact solutions of the VF-learning neural model (dashed line), the DVI-computed solutions of the VF-learning neural model with $h = T_{\text{DT}}/20$ (scatter dots) and the reference trajectories (gray line).
    \review{The legend only denotes the line or marker style, unrelated to the color.}
    }
    \label{fig:gc:VF_rz}
\end{figure}

\begin{figure}
    \centering
    \includegraphics[width=0.999\textwidth]{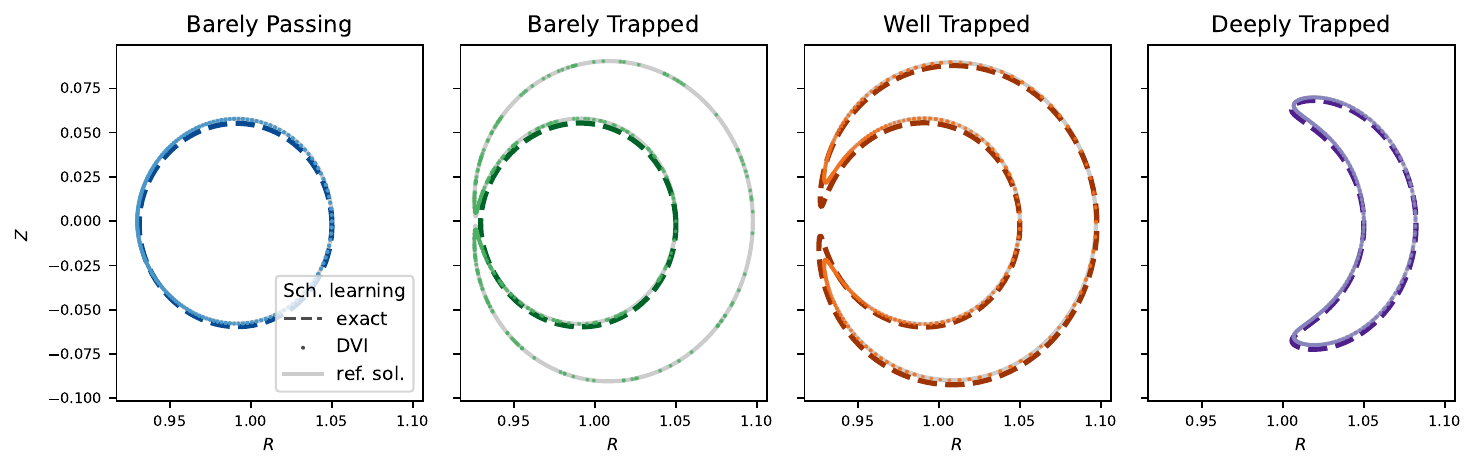}%
    \caption{(Guiding center - scheme learning) Comparisons of the exact solutions of the scheme-learning neural model (dashed line), the DVI-computed solutions of the scheme-learning neural model with $h = T_{\text{DT}}/20$ (scatter dots) and the reference trajectories (gray line).}
    \label{fig:gc:sch_rz}
\end{figure}

\begin{figure}
    \centering
    \includegraphics[width=0.999\textwidth]{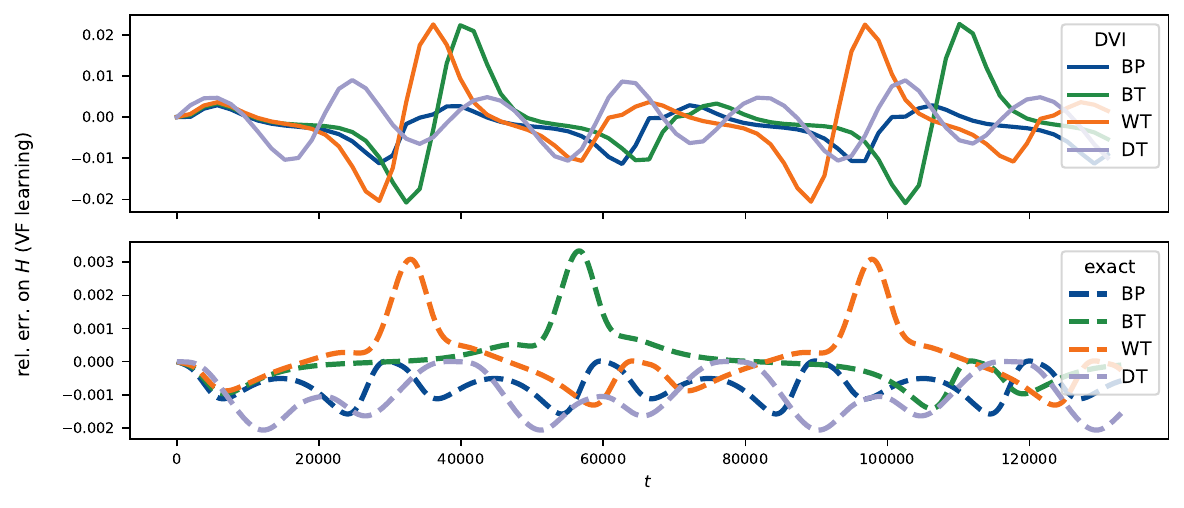}%
    \caption{(Guiding center - VF learning) Relative variations of the Hamiltonian~\eqref{eq:gc:Hamiltonian} for each trajectory of the VF-learning model using the DVI, with  $h = T_{\text{DT}}/20$, (top) and a refined simulation (bottom).}
    \label{fig:gc:VF_err_h}
\end{figure}

\begin{figure}
    \centering
    \includegraphics[width=0.999\textwidth]{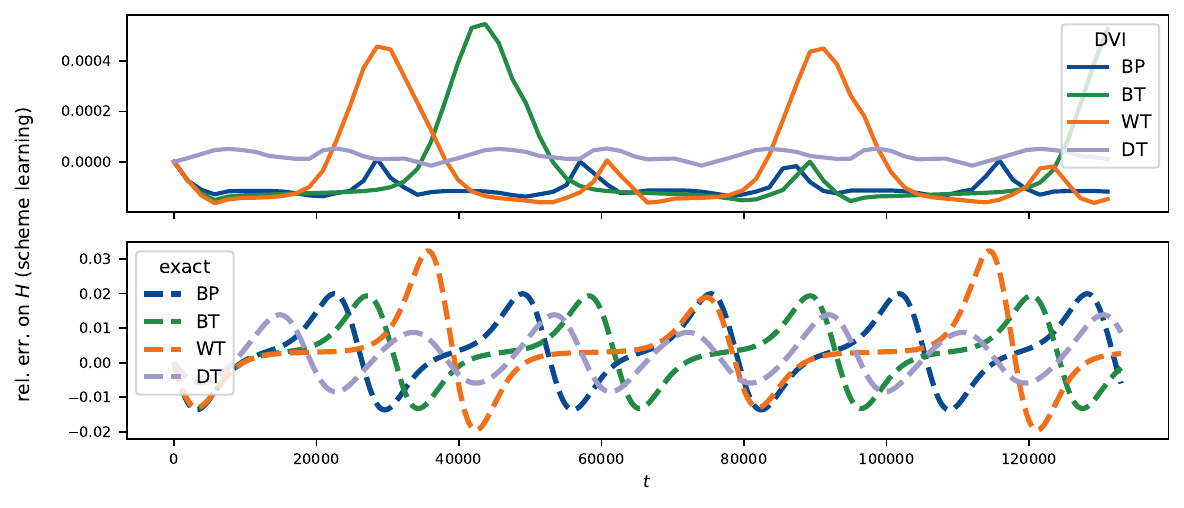}%
    \caption{(Guiding center - scheme learning) Relative variations of the Hamiltonian~\eqref{eq:gc:Hamiltonian} for each trajectory of the scheme-learning model using the DVI, with  $h = T_{\text{DT}}/20$, (top) and a refined simulation (bottom).}
    \label{fig:gc:sch_err_h}
\end{figure}

The observations made on the BP trajectory are confirmed in Figures~\ref{fig:gc:VF_rz} and~\ref{fig:gc:sch_rz}, where we compare the exact solutions (computed with \texttt{solve\_ivp} as in the other experiments), the DVI-computed numerical solutions, and the reference trajectory. 
The VF-learning model produced very accurate exact solutions (in dashed lines), while the scheme-learning model is very accurate in its numerical solution using the DVI (scatter dots).
Comparing the VF-learning model with the reference model (Figure~\ref{fig:charactRefGuidingCenter}), the former actually behaves better than the latter when using the DVI---especially with the BT trajectory, which remains trapped with the neural model (BP also remains passing).
The impact of correcting terms in scheme learning is clear with the BT trajectory, which has the wrong behavior when considering the exact solution, but is more accurate than VF learning when applying the DVI.
This is all confirmed in Figures~\ref{fig:gc:VF_err_h} and~\ref{fig:gc:sch_err_h}.
As in previous cases, this confirms that the network learns a very precise discrete flow, which effectively cancels or minimizes the scheme’s error terms. 
The scheme-learning model is a better choice for long-time simulations, as long as the scheme and time-steps match the training.

\begin{figure}
    \centering
    \includegraphics[width=0.999\textwidth]{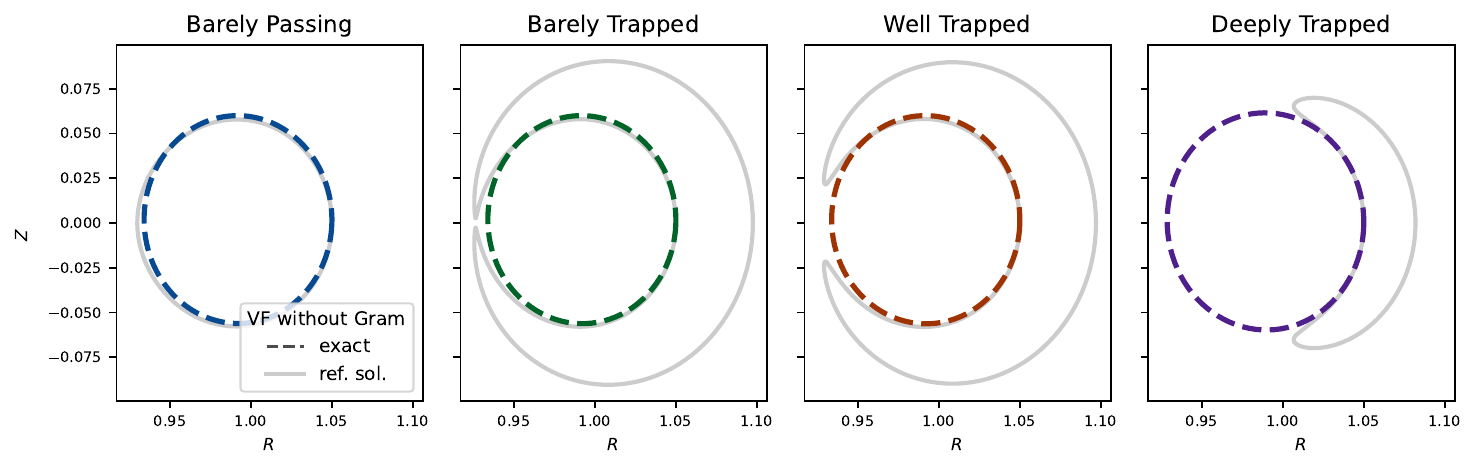}%
    \caption{(Guiding center - VF-learning without Gram-informed norm) Exact solutions of the VF-learning neural model (dashed line) trained without the multiscale norm~\eqref{eq:normVFErr}, compared to the reference trajectories (gray line).}
    \label{fig:gc:no_gram_rz}
\end{figure}

In Figure~\ref{fig:gc:no_gram_rz}, we illustrate the importance of using a multiscale norm (in our case, a Gram-informed norm). Without it, some components are completely ignored by the loss, since $|\dot u| \sim 10^{-4} \ll |\dot r| \sim 10^{-2} \ll |\dot \theta| \sim 1$. The trained model is then qualitatively inaccurate, and does not capture the diversity of trajectories displayed by the reference model.

\section{Conclusion}

In this paper, we build learning-based models to capture families of temporal trajectories, given by non-canonical Hamiltonian systems. We showed that, unlike the canonical case, one cannot directly apply a DVI scheme to a model trained with an imposed non-canonical structure. Indeed, non-canonical models are defined only up to a potential, whereas the DVI scheme is not invariant under such a potential shift. As a result, one may end up with a situation where the learning procedure captures the system correctly up to a potential, but injecting this model into the scheme leads to severe stability issues.

This highlights that, for more complex cases, one must not only impose the structure during training but also ensure compatibility between the learned model, its invariance properties, and the numerical scheme.

To address this issue, we proposed two approaches.
\begin{enumerate}
    \item learn the continuous-time model but add a penalty on the residual terms of the scheme, which mitigates the problem. However, the scheme remains of low order and becomes inaccurate for large time steps.
    \item learn using the scheme itself, which amounts to learning a discrete flow. This resolves the issue and improves accuracy (since it partly corrects the scheme’s consistency error), but the method is limited to the training time step.
\end{enumerate}

Our next objective is to tackle even more complex problems, such as Poisson systems. One could also enrich the learning procedure with loss functions that account for the integration of the entire trajectory, either using discrete approaches (unrolling) or continuous ones (Neural ODEs).

\backmatter

\section*{Declarations}

\bmhead{Conflict of Interest}
The authors declared they have no conflict of interests.

\bmhead{Funding}
This research was funded in part by l'Agence Nationale de la Recherche (ANR), project ANR-21-CE46-0014 (Milk).

\bmhead{Author Contributions}
CC supervised the project and provided feedback through proofreading. EF supervised the project, contributed references, wrote parts of the introduction and interpretation, and revised the manuscript. MK supervised the project, provided expertise on the numerical methods and test cases, and gave feedback through proofreading. LN supervised the project, contributed to visualizations, and carried out major revisions of the text. LT designed the training procedures, developed the code, performed the experiments, drafted the manuscript, and carried out major revisions.

\bmhead{Acknowledgement}
This research was funded in part by l'Agence Nationale de la Recherche (ANR), project ANR-21-CE46-0014 (Milk).

\bmhead{Supplementary information}

The code used to perform experiments can be found at \url{https://github.com/tremelow/symplearn/releases/tag/preprint}.

\begin{appendices}

\section{Using degenerate variational integrators}

\subsection{The Jacobian of variational integrators}

Usual variational integrators are based on discrete Lagrangians $(z_0, z_1) \mapsto L_h(z_0, z_1)$ and lead to the well-known discrete Euler-Lagrange equation
\begin{equation}
\label{eq:std-del}
    D_2 L_h(z_{n-1}, z_n) + D_1 L_h(z_n, z_{n+1}) = 0 .
\end{equation}
Many papers actually focus on learning a (continuous) Lagrangian from which they obtain discrete Euler-Lagrange equations which constitute the loss.
To avoid the trivial solution $L \rightarrow 0$, a non-triviality condition is required. 
A natural way to do this is to ensure that the Jacobian w.r.t. $z_{n+1}$ in~\eqref{eq:std-del}
\begin{equation}
    \label{eq:jac-std-del}
    J_{n+1}^{[{\rm DEL}]} = D_2 D_1 L_h (z_n, z_{n+1})
\end{equation}
is nicely invertible, such that Newton-Raphson iterations are well-defined. Let us mention two approaches which try to minimize the average value on the training dataset of the following quantities,
\begin{equation*}
\begin{cases}
    \text{\cite{Lishkova.Scherer.ea.2022.DiscreteLagrangianNeural}:} & \displaystyle\frac{1}{1 - e^{-0.01\: {\rm det}(J)}},
    \\ 
    \text{\cite{Offen.Ober-Blobaum.2024.LearningDiscreteModels}:} & \displaystyle \mathrm{ReLU} \left( 1 - \frac{10}{\| J^{-1} \|^2} \right) .
\end{cases}
\end{equation*}
Both kill two birds with one stone by penalising trivial Lagrangians and ensuring the invertibility of the Jacobian.
With the midpoint scheme $L_h(z_0, z_1) = L\left( \frac{z_0 + z_1}{2}, \frac{z_1 - z_0}{h} \right)$, this induces non-degenerate Lagrangian, since $D_2 D_1 L_h = \frac{1}{4} \partial^2_{z z} L - \frac{1}{h^2} \partial^2_{\dot z \dot z} L$.

In the case of degenerate Lagrangians, the principal term $\partial^2_{\dot z \dot z} L$ vanishes, therefore this reasoning must be adapted.
This is also apparent for the first-order DVI~\eqref{eq:Scheme}, as the DEL equation changes due to the difference between the $x$ and $y$ components.
It becomes
\begin{equation*}
    \left\{\begin{aligned}
        D_2 L_h(x_{n-1}, x_n, y_n) + D_1 L_h(x_n, x_{n+1}, y_{n+1}) = 0 , \\
        D_3 L_h(x_n, x_{n+1}, y_{n+1}) = 0 .
    \end{aligned}\right.
\end{equation*}
The Jacobian of the system w.r.t. $(x_{n+1}, y_{n+1})$ is no longer~\eqref{eq:jac-std-del} but a block matrix
\begin{equation*}
    J_{n+1}^{[{\rm DEL}]} = \begin{bmatrix}
        D_2 D_1 & D_3 D_1 \\ D_2 D_3 & D_3^{\ 2}
    \end{bmatrix} L_h (x_n, x_{n+1}, y_{n+1}) .
\end{equation*}
Explicitly, the blocks are
\begin{equation*}
\begin{aligned}
    D_2 D_1 L_h & = \left( -\frac{1}{h} \vartheta_{i,j} \right)_{i,j} \qquad&
    D_2 D_3 L_h & = \left( \vartheta_{k,\alpha,j} \cdot \frac{x_{n+1}^k - x_n^k}{h} + \frac{1}{h}\vartheta_{j,\alpha} - H_{,\alpha,j} \right)_{\alpha,j} \\
    D_3 D_1 L_h & = \left( -\frac{1}{h} \vartheta_{i,\beta} \right)_{i,\beta} &
    D_3^{\ 2} L_h & = \left( \vartheta_{k,\alpha,\beta} \cdot \frac{x_{n+1}^k - x_n^k}{h} - H_{,\alpha,\beta} \right)_{\alpha,\beta}
\end{aligned}
\end{equation*}

\subsection{Rigorous error estimate of the scheme}
\label{appendix:error}

Given initial conditions $x_0, y_0$ denote $x_h \approx x(h), y_h \approx y(h)$ the result of after one step of the scheme~\eqref{eq:Scheme}. 
We wish to find an expression of the leading term in the error $\begin{bmatrix} x(h) - x_h \\ y(h) - y_h \end{bmatrix}$. 
For this, we perform expansions, exploiting the assumed result $z_h = z_0 + h\dot z(0) + \mathcal{O}(h^2)$,
\begin{equation*}
    \vartheta_{j,\alpha}(z_h) = \vartheta_{j,\alpha} + h \vartheta_{j,\alpha,k} \dot z^k(0) + \mathcal{O}(h^2) , \qquad
    H_{,\alpha}(z_h) = H_{,\alpha} + h H_{,\alpha,k} \dot z^k(0) + \mathcal{O}(h^2) ,
\end{equation*}
where the maps $\vartheta_j, H$ and their differentials are evaluated at $z_0$. Therefore, the part of the scheme updating $x$ can be written
\begin{equation*}
    \bigl( \vartheta_{j,\alpha} + h \vartheta_{j,\alpha,k} \dot z^k + \mathcal{O}(h^2) \bigr) (x_h^j - x_0^j) = h H_{,\alpha} + h^2 H_{,\alpha,k} \dot z^k + \mathcal{O}(h^3) ,
\end{equation*}
where the maps and $\dot z$ are evaluated time $t = 0$. 
In the sequel, the argument of $\vartheta, H$ and their differentials is always $x_0$, and the time-differentials are always evaluated at time $t = 0$, so that we may lighten the presentation.
Using $x_h^j - x_j^0 = h \dot x^j + \mathcal{O}(h^2) = \mathcal{O}(h)$, we find
\begin{equation*}
    \vartheta_{j,\alpha} x_h^j = \vartheta_{j,\alpha} x_0^j + h H_{,\alpha} + h^2 \bigl( H_{,\alpha,k} - \vartheta_{j,\alpha,k} \dot x^j \bigr) \dot z^k + \mathcal{O}(h^3) .
\end{equation*}
On the other hand, $x(h) = x_0 + h \dot x + \frac12 h^2 \ddot x + \mathcal{O}(h^3)$, where the second-order derivative may be obtained from the differential equation~\eqref{eq:degenerate_vf}, as
\begin{equation*}
    0 = \vartheta_{j,\alpha} \ddot x^j + \vartheta_{j,\alpha,k} \dot x^j \dot z^k - H_{,\alpha,k} \dot z^k .
\end{equation*}
Therefore, with $\vartheta_{j,\alpha} \dot x^j = H_{,\alpha}$, the dominant error term is
\begin{equation*}
    \vartheta_{j,\alpha} \bigl( x^j(h) - x_h^j \bigr) = \frac{h^2}{2} \left( \vartheta_{j,\alpha,k} \dot x^j - H_{,\alpha,k} \right) \dot z^k  + \mathcal{O}(h^3),
\end{equation*}
i.e. in matrix-vector form, recognizing the Lagrangian,
\begin{equation}
    x(h) - x_h = -\frac{h^2}{2} \ddot x + \mathcal{O}(h^3) = \frac{h^2}{2} (D_y \vartheta)^{-\mathsf{T}} D_z \bigl[ \nabla_y L \bigr] \dot z + \mathcal{O}(h^3),
\end{equation}
where the differentials w.r.t. $y$ and $z$ are not applied to $\dot x$.

For the estimate in $y$, we recognize $x_{-h} = x_0 - h \dot x_{|t = 0}$, therefore the scheme is
\begin{equation*}
    \vartheta_i (x_h, y_h) = \vartheta_i + h \vartheta_{j,i} \dot x^j - h H_{,i} .
\end{equation*}
A straightforward Taylor expansion also yields 
\begin{equation*}
    \vartheta_i(x_h, y_h) = \vartheta_i + \vartheta_{i,j} (x_h^j - x_0^j) + \vartheta_{i,\nu} (y_h^\nu - y_0^\nu) + \frac{h^2}{2} \vartheta_{i,k,k'} \dot z^k \dot z^{k'} + \mathcal{O}(h^3) .
\end{equation*}
Reordering the terms, we find
\begin{equation*}
    \vartheta_{i,\nu} y_h^\nu = \vartheta_{i,\nu} y_0^\nu + h \vartheta_{j,i}\dot x^j - h\vartheta_{i,j} \frac{x^j_h - x^j_0}{h} - h H_{,i} - \frac{h^2}{2} \vartheta_{i,k,k'} \dot z^k \dot z^{k'} + \mathcal{O}(h^3)
\end{equation*}
Since $x_h = x_0 + h \dot x + h^2 \ddot x + \mathcal{O}(h^3)$, we find
\begin{equation*}
    \vartheta_{i,\nu} y_h^\nu = \vartheta_{i,\nu} y_0^\nu + h \underbrace{\bigl( (\vartheta_{j,i} - \vartheta_{i,j}) \dot x^j - H_{,i} \bigr)}_{= \vartheta_{i,\nu} \dot y^\nu} - h^2 \left( \vartheta_{i,j} \ddot x^j + {\textstyle\frac{1}{2}} \vartheta_{i,k,k'} \dot z^k \dot z^{k'} \right) + \mathcal{O}(h^3) .
\end{equation*}
The exact solution satisfies $y(h) = y_0 + h \dot y + \frac12 h^2 \ddot y + \mathcal{O}(h^3)$, therefore the error term is
\begin{equation}
\label{app:eq:error_term_y}
    \vartheta_{i,\nu} \left( y^\nu(h) - y^\nu_h \right) = \frac{h^2}{2} \vartheta_{i,\nu} \ddot y^\nu + h^2 \left( \vartheta_{i,j} \ddot x^j + {\textstyle\frac{1}{2}} \vartheta_{i,k,k'} \dot z^k \dot z^{k'} \right) + \mathcal{O}(h^3) .
\end{equation}
The second-order derivative on $y$ is obtained from
\begin{equation*}
    0 = \vartheta_{i,\nu} \ddot y^\nu + \vartheta_{i,\nu,k} \dot y^\nu \dot z^k + \bigl( \vartheta_{i,j} - \vartheta_{j,i} \bigr) \ddot x^j + \bigl( \vartheta_{i,j,k} - \vartheta_{j,i,k} \bigr) \dot x^j \dot z^k + H_{,i,k} \dot z^k ,
\end{equation*}
which, by combining $\vartheta_{i,j,k'}\dot x^j + \vartheta_{i,\nu,k'}\dot y^\nu = \vartheta_{i,k,k'}\dot z^k$, is reordered
\begin{equation*}
    \vartheta_{i,\nu} \ddot y^\nu = (\vartheta_{j,i} - \vartheta_{i,j}) \ddot x^j - \vartheta_{i,k,k'} \dot z^k \dot z^{k'} + (\vartheta_{j,i,k} \dot x^j - H_{,i,k}) \dot z^k .
\end{equation*}
This may finally be injected in the error term to find
\begin{equation*}
\begin{aligned}
    \vartheta_{i,\nu} \left( y^\nu(h) - y^\nu_h \right)
    & = \frac{h^2}{2} \left( (\vartheta_{j,i} - \vartheta_{i,j}) \ddot x^j - \vartheta_{i,k,k'} \dot z^k \dot z^{k'} + (\vartheta_{j,i,k} \dot x^j - H_{,i,k}) \dot z^k \right) \\
    & + h^2 \left( \vartheta_{i,j} \ddot x^j + \frac12 \vartheta_{i,k,k'} \dot z^k \dot z^{k'} \right) + \mathcal{O}(h^3) ,
\end{aligned}
\end{equation*}
After simplification, this is written in the vector form
\begin{equation}
    y(h) - y_h = \frac{h^2}{2} (D_y\vartheta)^{-1} \left( (D_x\vartheta + D_x\vartheta^\mathsf{T}) \ddot x + D_z [\nabla_x L ] \dot z \right) + \mathcal{O}(h^3) 
\end{equation}
where, again, the differentials w.r.t. $x$ and $z$ are only applied to the maps of the Lagrangian, not to $\dot x$.
The second-order derivative $\ddot x$ may also be replaced using the error term $x(h) - x_h$, which yields
\begin{equation}
    \begin{bmatrix}
    D_x\vartheta + D_x\vartheta^\mathsf{T} & D_y\vartheta \\
    D_y\vartheta^\mathsf{T} & 0
    \end{bmatrix} \begin{bmatrix} x(h) - x_h \\ y(h) - y_h \end{bmatrix}
    = \frac{h^2}{2} D_z [\nabla_z L] \dot z + \mathcal{O}(h^3) .
\end{equation}

\begin{remark}
\label{app:rmk:gauge_choice}
    Noticing the identity $\vartheta_{i,j} \ddot x^j + \frac12 \vartheta_{i,k,k'} \dot z^k \dot z^{k'} = \frac12 \left(\ddot \vartheta + D_x\vartheta \ddot x - D_y\vartheta \ddot y \right)$, the error term in~\eqref{app:eq:error_term_y} is simply
    \begin{equation*}
        D_y\vartheta \left( y(h) - y_h \right) = \frac{h^2}{2} \left( \ddot\vartheta + D_x\vartheta\ \ddot x \right) + \mathcal{O}(h^3) ,
    \end{equation*}
    i.e., the regularity of $\vartheta$ impacts the accuracy of the scheme.
    Similarly, the symmetric part of $D_x \vartheta$ appears in this error term, even though has no influence on the original dynamics.
    All this explains why the gauge choice matters in this context.
\end{remark}

\section{Brief study of the guiding center}
\subsection{Deriving the dynamics}

It is standard to write these equations using the \textit{effective} magnetic potential $A_\phi^\dagger := \vartheta_\phi = A_\phi + u$.
Assuming only $A_{\theta,\phi} = A_{\theta,u} = 0$, the Euler-Lagrange equations are 
\begin{equation*}
    \begin{aligned}
        A_{\theta,r}\dot r + A_{\theta,\theta}\dot\theta &= A_{\theta,\theta} \dot\theta + A_{\phi,\theta}^\dagger\dot\phi - H_{,\theta} \\
        A_{\phi,r}^\dagger\dot r + A_{\phi,\theta}^\dagger\dot\theta + A_{\phi,\phi}^\dagger\dot\phi + A_{\phi,u}^\dagger\dot u &= A_{\phi,\phi}^\dagger\dot\phi - H_{,\phi} \\
        0 &= A_{\theta,r}\dot\theta + A_{\phi,r}^\dagger\dot\phi - H_{,r} \\
        0 &= A_{\phi,u}^\dagger\dot\phi - H_{,u} .
    \end{aligned}
\end{equation*}
Considering the lines in order $(4,3,1,2)$, this can be transformed into a triangular system,
\begin{equation*}
    \dot\phi = \frac{H_{,u}}{A_{\phi,u}^\dagger}, \quad 
    \dot\theta = \frac{H_{,r} - A_{\phi,r}^\dagger\dot\phi}{A_{\theta,r}}, \quad 
    \dot r = \frac{A_{\phi,\theta}^\dagger\dot\phi - H_{,\theta}}{A_{\theta,r}}, \quad
    \dot u = -\frac{A_{\phi,\theta}^\dagger\dot\theta + A_{\phi,r}^\dagger\dot r + H_{,\phi}}{A_{\phi,u}^\dagger} .
\end{equation*}
These dynamics are well defined only if $A_{\theta,r}$ and $A_{\phi,u}^\dagger$ are non-zero, which corresponds to the invertibility of $D_y \vartheta = \begin{pmatrix} A_{\theta,r} & 0 \\ A_{\phi,r} & A_{\phi,u}^\dagger \end{pmatrix}$. 
Reordering into $(\theta,\phi,r,u)$, injecting the expression of the Hamiltonian and of $A_\phi^\dagger = A_\phi + u$, along with the assumption $A_{\phi,\theta} = A_{\phi,\phi} = 0$, finally yields 
\begin{equation*}
    \dot\theta = \frac{\mu B_{,r} - A_{\phi,r}u}{A_{\theta,r}}, \qquad
    \dot\phi = u, \qquad
    \dot r = -\frac{\mu B_{,\theta}}{A_{\theta,r}}, \qquad
    \dot u = \frac{\mu B_{,\theta} A_{\phi,r}}{A_{\theta,r}}, 
\end{equation*}
Denoting $\rho = r\cos(\theta)/R_0$, $Z = r\sin(\theta)/R_0$, $\widetilde{r} = r/(q_0 R_0)$, these quantities are
\begin{equation}
\label{eq:derivsMagGC}
\begin{array}{c}\displaystyle
    A_{\theta,r} = \frac{B_0 r}{1 + \rho}, \qquad 
    A_{\phi,r} = -\frac{B_0 r}{q_0} , \qquad
    B_{,r} = \frac{B_0\sqrt{1 + \widetilde{r}^2}}{r(1 + \rho)} \left( \frac{\widetilde{r}^2}{1 + \widetilde{r}^2} - \frac{\rho}{1 + \rho} \right) ,
\vspace{\medskipamount}\\ \displaystyle
    A_{\theta,\theta} = \frac{B_0 r^2 Z}{\rho^3} \left(2\log(1 + \rho) - 2\rho + \frac{\rho^2}{1 + \rho} \right), \qquad 
    B_{,\theta} = \frac{B_0 Z \sqrt{1 + \widetilde{r}^2}}{(1 + \rho)^2} .
\end{array}
\end{equation}

\subsection{Computing the magnetic potential and its differentials}

The expressions of $A_\theta$ and $A_{\theta,\theta}$ are unstable for small values of $\cos(\theta)$, even though the functions are well-defined in this limit.
Therefore, we compute them using an integral form, 
\begin{equation*}
    A_{\theta}(\theta,r) = B_0 r^2 \int_0^1 \frac{t\: {\rm d}t}{1 + \rho t}, \qquad
    A_{\theta,\theta}(\theta,r) = B_0 r^2 Z \int_0^1 \frac{t^2\:{\rm d}t}{\left( 1 + \rho t \right)^2} .
\end{equation*}
where, again, $\rho = r\cos(\theta)/R_0$ and $Z = r\sin(\theta)/R_0$.
This is more costly to evaluate and less accurate for $\rho \approx -1$, but uniformly stable for all values of $\theta$.

We compute $A_\theta$ and $A_{\theta,\theta}$ using this form and evaluating the integral with a Gauss-Legendre integration. 
A highly-accurate reference value was computed in \texttt{Julia} using \texttt{BigFloat} and 50 collocation points. 
Using double-precision with 20 collocation points, this method computes the integral in $A_{\theta,\theta}$ with an error of $1.7\cdot 10^{-11}$ for $\rho = -0.9$, and of less than $10^{-15}$ for $-0.8 \leq \rho \leq 0.9$. 
The error is slightly smaller for the integral in $A_\theta$.
For comparison, using $\rho = 10^{-6}$ in the closed formula from~\eqref{eq:derivsMagGC} yields an error of $2.1\cdot 10^{-4}$ with double-precision floats. 

Some floating-point magic tricks might make the closed formula more accurate, using e.g. \texttt{log1p},\footnote{%
    For example, computing $\bigl({\tt log1p}(x)/x - 1 \bigr) / \bigl((1+x) - 1 \bigr)$ is more accurate than $\bigl({\tt log1p}(x)/x - 1 \bigr) / x$ for small values of $x$. 
} 
but we did not find any, and they might be processor-dependent.

\end{appendices}


\bibliography{geoml}


\begin{thebibliography}{42}
\ifx \bisbn   \undefined \def \bisbn  #1{ISBN #1}\fi
\ifx \binits  \undefined \def \binits#1{#1}\fi
\ifx \bauthor  \undefined \def \bauthor#1{#1}\fi
\ifx \batitle  \undefined \def \batitle#1{#1}\fi
\ifx \bjtitle  \undefined \def \bjtitle#1{#1}\fi
\ifx \bvolume  \undefined \def \bvolume#1{\textbf{#1}}\fi
\ifx \byear  \undefined \def \byear#1{#1}\fi
\ifx \bissue  \undefined \def \bissue#1{#1}\fi
\ifx \bfpage  \undefined \def \bfpage#1{#1}\fi
\ifx \blpage  \undefined \def \blpage #1{#1}\fi
\ifx \burl  \undefined \def \burl#1{\textsf{#1}}\fi
\ifx \doiurl  \undefined \def \doiurl#1{\url{https://doi.org/#1}}\fi
\ifx \betal  \undefined \def \betal{\textit{et al.}}\fi
\ifx \binstitute  \undefined \def \binstitute#1{#1}\fi
\ifx \binstitutionaled  \undefined \def \binstitutionaled#1{#1}\fi
\ifx \bctitle  \undefined \def \bctitle#1{#1}\fi
\ifx \beditor  \undefined \def \beditor#1{#1}\fi
\ifx \bpublisher  \undefined \def \bpublisher#1{#1}\fi
\ifx \bbtitle  \undefined \def \bbtitle#1{#1}\fi
\ifx \bedition  \undefined \def \bedition#1{#1}\fi
\ifx \bseriesno  \undefined \def \bseriesno#1{#1}\fi
\ifx \blocation  \undefined \def \blocation#1{#1}\fi
\ifx \bsertitle  \undefined \def \bsertitle#1{#1}\fi
\ifx \bsnm \undefined \def \bsnm#1{#1}\fi
\ifx \bsuffix \undefined \def \bsuffix#1{#1}\fi
\ifx \bparticle \undefined \def \bparticle#1{#1}\fi
\ifx \barticle \undefined \def \barticle#1{#1}\fi
\bibcommenthead
\ifx \bconfdate \undefined \def \bconfdate #1{#1}\fi
\ifx \botherref \undefined \def \botherref #1{#1}\fi
\ifx \url \undefined \def \url#1{\textsf{#1}}\fi
\ifx \bchapter \undefined \def \bchapter#1{#1}\fi
\ifx \bbook \undefined \def \bbook#1{#1}\fi
\ifx \bcomment \undefined \def \bcomment#1{#1}\fi
\ifx \oauthor \undefined \def \oauthor#1{#1}\fi
\ifx \citeauthoryear \undefined \def \citeauthoryear#1{#1}\fi
\ifx \endbibitem  \undefined \def \endbibitem {}\fi
\ifx \bconflocation  \undefined \def \bconflocation#1{#1}\fi
\ifx \arxivurl  \undefined \def \arxivurl#1{\textsf{#1}}\fi
\csname PreBibitemsHook\endcsname

\bibitem[\protect\citeauthoryear{Ardizzone
  et~al.}{2019}]{Ardizzone.Kruse.ea.2019.AnalyzingInverseProblems}
\begin{botherref}
\oauthor{\bsnm{Ardizzone}, \binits{L.}},
\oauthor{\bsnm{Kruse}, \binits{J.}},
\oauthor{\bsnm{Wirkert}, \binits{S.}},
\oauthor{\bsnm{Rahner}, \binits{D.}},
\oauthor{\bsnm{Pellegrini}, \binits{E.W.}},
\oauthor{\bsnm{Klessen}, \binits{R.S.}},
\oauthor{\bsnm{{Maier-Hein}}, \binits{L.}},
\oauthor{\bsnm{Rother}, \binits{C.}},
\oauthor{\bsnm{K{\"o}the}, \binits{U.}}:
Analyzing {{Inverse Problems}} with {{Invertible Neural Networks}}.
arXiv
(2019)
\end{botherref}
\endbibitem

\bibitem[\protect\citeauthoryear{Bouchereau
  et~al.}{2023}]{Bouchereau.Chartier.ea.2023.MachineLearningMethodsa}
\begin{botherref}
\oauthor{\bsnm{Bouchereau}, \binits{M.}},
\oauthor{\bsnm{Chartier}, \binits{P.}},
\oauthor{\bsnm{Lemou}, \binits{M.}},
\oauthor{\bsnm{M{\'e}hats}, \binits{F.}}:
Machine {{Learning Methods}} for {{Autonomous Ordinary Differential
  Equations}}.
arXiv
(2023).
\doiurl{10.48550/arXiv.2304.09036}
\end{botherref}
\endbibitem

\bibitem[\protect\citeauthoryear{Burby
  et~al.}{2022}]{Burby.Finn.ea.2022.ImprovedAccuracyDegenerate}
\begin{barticle}
\bauthor{\bsnm{Burby}, \binits{J.W.}},
\bauthor{\bsnm{Finn}, \binits{J.M.}},
\bauthor{\bsnm{Ellison}, \binits{C.L.}}:
\batitle{Improved accuracy in degenerate variational integrators for guiding
  centre and magnetic field line flow}.
\bjtitle{J. Plasma Phys.}
\bvolume{88}(\bissue{2}),
\bfpage{835880201}
(\byear{2022})
\doiurl{10.1017/S0022377821001136}
\end{barticle}
\endbibitem

\bibitem[\protect\citeauthoryear{Bonneville
  et~al.}{2024}]{bonneville2024comprehensive}
\begin{botherref}
\oauthor{\bsnm{Bonneville}, \binits{C.}},
\oauthor{\bsnm{He}, \binits{X.}},
\oauthor{\bsnm{Tran}, \binits{A.}},
\oauthor{\bsnm{Park}, \binits{J.S.}},
\oauthor{\bsnm{Fries}, \binits{W.}},
\oauthor{\bsnm{Messenger}, \binits{D.A.}},
\oauthor{\bsnm{Cheung}, \binits{S.W.}},
\oauthor{\bsnm{Shin}, \binits{Y.}},
\oauthor{\bsnm{Bortz}, \binits{D.M.}},
\oauthor{\bsnm{Ghosh}, \binits{D.}},
\oauthor{\bsnm{Chen}, \binits{J.-S.}},
\oauthor{\bsnm{Belof}, \binits{J.}},
\oauthor{\bsnm{Choi}, \binits{Y.}}:
A comprehensive review of latent space dynamics identification algorithms for
  intrusive and non-intrusive reduced-order-modeling.
arXiv preprint arXiv:2403.10748
(2024)
\end{botherref}
\endbibitem

\bibitem[\protect\citeauthoryear{Brunton et~al.}{2016}]{brunton2016discovering}
\begin{barticle}
\bauthor{\bsnm{Brunton}, \binits{S.L.}},
\bauthor{\bsnm{Proctor}, \binits{J.L.}},
\bauthor{\bsnm{Kutz}, \binits{J.N.}}:
\batitle{Discovering governing equations from data by sparse identification of
  nonlinear dynamical systems}.
\bjtitle{PNAS}
\bvolume{113}(\bissue{15}),
\bfpage{3932}--\blpage{3937}
(\byear{2016})
\end{barticle}
\endbibitem

\bibitem[\protect\citeauthoryear{Blondel and
  Roulet}{2024}]{blondel2024elements}
\begin{botherref}
\oauthor{\bsnm{Blondel}, \binits{M.}},
\oauthor{\bsnm{Roulet}, \binits{V.}}:
The elements of differentiable programming.
arXiv preprint arXiv:2403.14606
(2024)
\end{botherref}
\endbibitem

\bibitem[\protect\citeauthoryear{Bilo{\v{s}} et~al.}{2021}]{bilovs2021neural}
\begin{barticle}
\bauthor{\bsnm{Bilo{\v{s}}}, \binits{M.}},
\bauthor{\bsnm{Sommer}, \binits{J.}},
\bauthor{\bsnm{Rangapuram}, \binits{S.S.}},
\bauthor{\bsnm{Januschowski}, \binits{T.}},
\bauthor{\bsnm{G{\"u}nnemann}, \binits{S.}}:
\batitle{Neural flows: Efficient alternative to neural odes}.
\bjtitle{Adv. Neural Inf. Process. Syst.}
\bvolume{34},
\bfpage{21325}--\blpage{21337}
(\byear{2021})
\end{barticle}
\endbibitem

\bibitem[\protect\citeauthoryear{Burby
  et~al.}{2021}]{Burby.Tang.ea.2021.FastNeuralPoincare}
\begin{barticle}
\bauthor{\bsnm{Burby}, \binits{J.W.}},
\bauthor{\bsnm{Tang}, \binits{Q.}},
\bauthor{\bsnm{Maulik}, \binits{R.}}:
\batitle{Fast neural {{Poincar{\'e}}} maps for toroidal magnetic fields}.
\bjtitle{Plasma Phys. Controlled Fusion}
\bvolume{63}(\bissue{2}),
\bfpage{024001}
(\byear{2021})
\doiurl{10.1088/1361-6587/abcbaa}
\end{barticle}
\endbibitem

\bibitem[\protect\citeauthoryear{C{\^o}te et~al.}{2024}]{cote2023hamiltonian}
\begin{barticle}
\bauthor{\bsnm{C{\^o}te}, \binits{R.}},
\bauthor{\bsnm{Franck}, \binits{E.}},
\bauthor{\bsnm{Navoret}, \binits{L.}},
\bauthor{\bsnm{Steimer}, \binits{G.}},
\bauthor{\bsnm{Vigon}, \binits{V.}}:
\batitle{Hamiltonian reduction using a convolutional auto-encoder coupled to an
  hamiltonian neural network}.
\bjtitle{Commun. Comput. Phys.}
\bvolume{37},
\bfpage{315}--\blpage{352}
(\byear{2024})
\end{barticle}
\endbibitem

\bibitem[\protect\citeauthoryear{C{\^o}te et~al.}{2025}]{franck2025reduced}
\begin{botherref}
\oauthor{\bsnm{C{\^o}te}, \binits{R.}},
\oauthor{\bsnm{Franck}, \binits{E.}},
\oauthor{\bsnm{Navoret}, \binits{L.}},
\oauthor{\bsnm{Steimer}, \binits{G.}},
\oauthor{\bsnm{Vigon}, \binits{V.}}:
Reduced particle in cell method for the vlasov-poisson system using
  auto-encoder and hamiltonian neural.
arXiv preprint arXiv:2506.15203
(2025)
\end{botherref}
\endbibitem

\bibitem[\protect\citeauthoryear{Cranmer
  et~al.}{2020}]{Cranmer.Greydanus.ea.2020.LagrangianNeuralNetworks}
\begin{botherref}
\oauthor{\bsnm{Cranmer}, \binits{M.}},
\oauthor{\bsnm{Greydanus}, \binits{S.}},
\oauthor{\bsnm{Hoyer}, \binits{S.}},
\oauthor{\bsnm{Battaglia}, \binits{P.}},
\oauthor{\bsnm{Spergel}, \binits{D.}},
\oauthor{\bsnm{Ho}, \binits{S.}}:
Lagrangian {{Neural Networks}}.
arXiv
(2020).
\doiurl{10.48550/arXiv.2003.04630}
\end{botherref}
\endbibitem

\bibitem[\protect\citeauthoryear{Chen
  et~al.}{2021}]{Chen.Matsubara.ea.2021.NeuralSymplecticForm}
\begin{bchapter}
\bauthor{\bsnm{Chen}, \binits{Y.}},
\bauthor{\bsnm{Matsubara}, \binits{T.}},
\bauthor{\bsnm{Yaguchi}, \binits{T.}}:
\bctitle{Neural {{Symplectic Form}}: {{Learning Hamiltonian Equations}} on
  {{General Coordinate Systems}}}.
In: \bbtitle{Adv. Neural Inf. Process. Syst.},
vol. \bseriesno{34},
pp. \bfpage{16659}--\blpage{16670}.
\bpublisher{Curran Associates, Inc.}, \blocation{???}
(\byear{2021})
\end{bchapter}
\endbibitem

\bibitem[\protect\citeauthoryear{Chen
  et~al.}{2018}]{Chen.Rubanova.ea.2018.NeuralOrdinaryDifferential}
\begin{bchapter}
\bauthor{\bsnm{Chen}, \binits{R.T.Q.}},
\bauthor{\bsnm{Rubanova}, \binits{Y.}},
\bauthor{\bsnm{Bettencourt}, \binits{J.}},
\bauthor{\bsnm{Duvenaud}, \binits{D.K.}}:
\bctitle{Neural {{Ordinary Differential Equations}}}.
In: \bbtitle{Adv. Neural Inf. Process. Syst.},
vol. \bseriesno{31}.
\bpublisher{Curran Associates, Inc.}, \blocation{???}
(\byear{2018})
\end{bchapter}
\endbibitem

\bibitem[\protect\citeauthoryear{Duruisseaux
  et~al.}{2023}]{Duruisseaux.Burby.ea.2023.ApproximationNearlyperiodicSymplectica}
\begin{barticle}
\bauthor{\bsnm{Duruisseaux}, \binits{V.}},
\bauthor{\bsnm{Burby}, \binits{J.W.}},
\bauthor{\bsnm{Tang}, \binits{Q.}}:
\batitle{Approximation of nearly-periodic symplectic maps via
  structure-preserving neural networks}.
\bjtitle{Sci. Rep.}
\bvolume{13}(\bissue{1}),
\bfpage{8351}
(\byear{2023})
\doiurl{10.1038/s41598-023-34862-w}
\end{barticle}
\endbibitem

\bibitem[\protect\citeauthoryear{Dupont et~al.}{2019}]{dupont2019augmented}
\begin{botherref}
\oauthor{\bsnm{Dupont}, \binits{E.}},
\oauthor{\bsnm{Doucet}, \binits{A.}},
\oauthor{\bsnm{Teh}, \binits{Y.W.}}:
Augmented neural odes.
Adv. Neural Inf. Process. Syst.
\textbf{32}
(2019)
\end{botherref}
\endbibitem

\bibitem[\protect\citeauthoryear{David and
  M{\'e}hats}{2021}]{David.Mehats.2021.SymplecticLearningHamiltonian}
\begin{botherref}
\oauthor{\bsnm{David}, \binits{M.}},
\oauthor{\bsnm{M{\'e}hats}, \binits{F.}}:
Symplectic learning for {{Hamiltonian Neural Networks}}.
arXiv
(2021).
\doiurl{10.48550/arXiv.2106.11753}
\end{botherref}
\endbibitem

\bibitem[\protect\citeauthoryear{Dong and
  Ni}{2021}]{Dong.Ni.2021.MethodRepresentingPeriodic}
\begin{barticle}
\bauthor{\bsnm{Dong}, \binits{S.}},
\bauthor{\bsnm{Ni}, \binits{N.}}:
\batitle{A method for representing periodic functions and enforcing exactly
  periodic boundary conditions with deep neural networks}.
\bjtitle{J. Comput. Phys.}
\bvolume{435},
\bfpage{110242}
(\byear{2021})
\doiurl{10.1016/j.jcp.2021.110242}
\end{barticle}
\endbibitem

\bibitem[\protect\citeauthoryear{Ellison
  et~al.}{2018}]{Ellison.Finn.ea.2018.DegenerateVariationalIntegrators}
\begin{barticle}
\bauthor{\bsnm{Ellison}, \binits{C.L.}},
\bauthor{\bsnm{Finn}, \binits{J.M.}},
\bauthor{\bsnm{Burby}, \binits{J.W.}},
\bauthor{\bsnm{Kraus}, \binits{M.}},
\bauthor{\bsnm{Qin}, \binits{H.}},
\bauthor{\bsnm{Tang}, \binits{W.M.}}:
\batitle{Degenerate variational integrators for magnetic field line flow and
  guiding center trajectories}.
\bjtitle{Phys. Plasma}
\bvolume{25}(\bissue{5}),
\bfpage{052502}
(\byear{2018})
\doiurl{10.1063/1.5022277}
\end{barticle}
\endbibitem

\bibitem[\protect\citeauthoryear{Eidnes and
  Lye}{2024}]{Eidnes.Lye.2024.PseudoHamiltonianNeuralNetworks}
\begin{barticle}
\bauthor{\bsnm{Eidnes}, \binits{S.}},
\bauthor{\bsnm{Lye}, \binits{K.O.}}:
\batitle{Pseudo-{{Hamiltonian}} neural networks for learning partial
  differential equations}.
\bjtitle{J. Comput. Phys.}
\bvolume{500},
\bfpage{112738}
(\byear{2024})
\doiurl{10.1016/j.jcp.2023.112738}
\end{barticle}
\endbibitem

\bibitem[\protect\citeauthoryear{Eidnes et~al.}{2023}]{eidnes2023pseudo}
\begin{barticle}
\bauthor{\bsnm{Eidnes}, \binits{S.}},
\bauthor{\bsnm{Stasik}, \binits{A.J.}},
\bauthor{\bsnm{Sterud}, \binits{C.}},
\bauthor{\bsnm{B{\o}hn}, \binits{E.}},
\bauthor{\bsnm{Riemer-S{\o}rensen}, \binits{S.}}:
\batitle{Pseudo-hamiltonian neural networks with state-dependent external
  forces}.
\bjtitle{Physica D: Nonlinear Phenomena}
\bvolume{446},
\bfpage{133673}
(\byear{2023})
\end{barticle}
\endbibitem

\bibitem[\protect\citeauthoryear{Farenga et~al.}{2025}]{farenga2025latent}
\begin{botherref}
\oauthor{\bsnm{Farenga}, \binits{N.}},
\oauthor{\bsnm{Fresca}, \binits{S.}},
\oauthor{\bsnm{Brivio}, \binits{S.}},
\oauthor{\bsnm{Manzoni}, \binits{A.}}:
On latent dynamics learning in nonlinear reduced order modeling.
Neural Networks,
107146
(2025)
\end{botherref}
\endbibitem

\bibitem[\protect\citeauthoryear{Greydanus
  et~al.}{2019}]{Greydanus.Dzamba.ea.2019.HamiltonianNeuralNetworks}
\begin{bchapter}
\bauthor{\bsnm{Greydanus}, \binits{S.}},
\bauthor{\bsnm{Dzamba}, \binits{M.}},
\bauthor{\bsnm{Yosinski}, \binits{J.}}:
\bctitle{Hamiltonian {{Neural Networks}}}.
In: \bbtitle{Adv. Neural Inf. Process. Syst.},
vol. \bseriesno{32}.
\bpublisher{Curran Associates, Inc.}, \blocation{???}
(\byear{2019})
\end{bchapter}
\endbibitem

\bibitem[\protect\citeauthoryear{Gnanasambandam
  et~al.}{2023}]{Gnanasambandam.Shen.ea.2023.Stanh}
\begin{botherref}
\oauthor{\bsnm{Gnanasambandam}, \binits{R.}},
\oauthor{\bsnm{Shen}, \binits{B.}},
\oauthor{\bsnm{Chung}, \binits{J.}},
\oauthor{\bsnm{Yue}, \binits{X.}},
\oauthor{\bsnm{Kong}, \binits{Z.}}:
Self-scalable tanh (stan): Multi-scale solutions for physics-informed neural
  networks.
IEEE Trans. Pattern Anal. Mach. Intell.
(2023)
\end{botherref}
\endbibitem

\bibitem[\protect\citeauthoryear{Hairer
  et~al.}{2006}]{Hairer.Lubich.ea.2006.GeometricNumericalIntegration}
\begin{bbook}
\bauthor{\bsnm{Hairer}, \binits{E.}},
\bauthor{\bsnm{Lubich}, \binits{C.}},
\bauthor{\bsnm{Wanner}, \binits{G.}}:
\bbtitle{Geometric {{Numerical}} {{Integration}}: {{Structure-Preserving}}
  {{Algorithms}} for {{Ordinary}} {{Differential}} {{Equations}}},
\bedition{2}nd edn.
\bsertitle{Springer {{Series}} in {{Computational Mathematics}}}.
\bpublisher{Springer},
\blocation{Berlin Heidelberg}
(\byear{2006}).
\doiurl{10.1007/3-540-30666-8}
\end{bbook}
\endbibitem

\bibitem[\protect\citeauthoryear{Jacot et~al.}{2018}]{jacot2018neural}
\begin{botherref}
\oauthor{\bsnm{Jacot}, \binits{A.}},
\oauthor{\bsnm{Gabriel}, \binits{F.}},
\oauthor{\bsnm{Hongler}, \binits{C.}}:
Neural tangent kernel: Convergence and generalization in neural networks.
Adv. Neural Inf. Process. Syst.
\textbf{31}
(2018)
\end{botherref}
\endbibitem

\bibitem[\protect\citeauthoryear{Jin
  et~al.}{2020}]{Jin.Zhang.ea.2020.LearningPoissonSystems}
\begin{botherref}
\oauthor{\bsnm{Jin}, \binits{P.}},
\oauthor{\bsnm{Zhang}, \binits{Z.}},
\oauthor{\bsnm{Kevrekidis}, \binits{I.G.}},
\oauthor{\bsnm{Karniadakis}, \binits{G.E.}}:
Learning {{Poisson}} Systems and Trajectories of Autonomous Systems via
  {{Poisson}} Neural Networks.
arXiv
(2020)
\end{botherref}
\endbibitem

\bibitem[\protect\citeauthoryear{Jin
  et~al.}{2020}]{Jin.Zhang.ea.2020.SympNetsIntrinsicStructurepreserving}
\begin{barticle}
\bauthor{\bsnm{Jin}, \binits{P.}},
\bauthor{\bsnm{Zhang}, \binits{Z.}},
\bauthor{\bsnm{Zhu}, \binits{A.}},
\bauthor{\bsnm{Tang}, \binits{Y.}},
\bauthor{\bsnm{Karniadakis}, \binits{G.E.}}:
\batitle{{{SympNets}}: {{Intrinsic}} structure-preserving symplectic networks
  for identifying {{Hamiltonian}} systems}.
\bjtitle{Neural Networks}
\bvolume{132},
\bfpage{166}--\blpage{179}
(\byear{2020})
\doiurl{10.1016/j.neunet.2020.08.017}
\end{barticle}
\endbibitem

\bibitem[\protect\citeauthoryear{Kaiser et~al.}{2018}]{kaiser2018sparse}
\begin{barticle}
\bauthor{\bsnm{Kaiser}, \binits{E.}},
\bauthor{\bsnm{Kutz}, \binits{J.N.}},
\bauthor{\bsnm{Brunton}, \binits{S.L.}}:
\batitle{Sparse identification of nonlinear dynamics for model predictive
  control in the low-data limit}.
\bjtitle{Proc. R. Soc. A}
\bvolume{474}(\bissue{2219}),
\bfpage{20180335}
(\byear{2018})
\end{barticle}
\endbibitem

\bibitem[\protect\citeauthoryear{Kolter and Manek}{2019}]{kolter2019learning}
\begin{botherref}
\oauthor{\bsnm{Kolter}, \binits{J.Z.}},
\oauthor{\bsnm{Manek}, \binits{G.}}:
Learning stable deep dynamics models.
Adv. Neural Inf. Process. Syst.
\textbf{32}
(2019)
\end{botherref}
\endbibitem

\bibitem[\protect\citeauthoryear{Lishkova
  et~al.}{2022}]{Lishkova.Scherer.ea.2022.DiscreteLagrangianNeural}
\begin{botherref}
\oauthor{\bsnm{Lishkova}, \binits{Y.}},
\oauthor{\bsnm{Scherer}, \binits{P.}},
\oauthor{\bsnm{Ridderbusch}, \binits{S.}},
\oauthor{\bsnm{Jamnik}, \binits{M.}},
\oauthor{\bsnm{Li{\`o}}, \binits{P.}},
\oauthor{\bsnm{{Ober-Bl{\"o}baum}}, \binits{S.}},
\oauthor{\bsnm{Offen}, \binits{C.}}:
Discrete {{Lagrangian Neural Networks}} with {{Automatic Symmetry Discovery}}.
arXiv
(2022)
\end{botherref}
\endbibitem

\bibitem[\protect\citeauthoryear{Lee et~al.}{2022}]{lee2022structure}
\begin{bchapter}
\bauthor{\bsnm{Lee}, \binits{K.}},
\bauthor{\bsnm{Trask}, \binits{N.}},
\bauthor{\bsnm{Stinis}, \binits{P.}}:
\bctitle{Structure-preserving sparse identification of nonlinear dynamics for
  data-driven modeling}.
In: \bbtitle{Mathematical and Scientific Machine Learning},
pp. \bfpage{65}--\blpage{80}
(\byear{2022}).
\bcomment{PMLR}
\end{bchapter}
\endbibitem

\bibitem[\protect\citeauthoryear{Massaroli
  et~al.}{2020}]{massaroli2020dissecting}
\begin{barticle}
\bauthor{\bsnm{Massaroli}, \binits{S.}},
\bauthor{\bsnm{Poli}, \binits{M.}},
\bauthor{\bsnm{Park}, \binits{J.}},
\bauthor{\bsnm{Yamashita}, \binits{A.}},
\bauthor{\bsnm{Asama}, \binits{H.}}:
\batitle{Dissecting neural odes}.
\bjtitle{Adv. Neural Inf. Process. Syst.}
\bvolume{33},
\bfpage{3952}--\blpage{3963}
(\byear{2020})
\end{barticle}
\endbibitem

\bibitem[\protect\citeauthoryear{Mathiesen
  et~al.}{2022}]{Mathiesen.Yang.ea.2022.HyperverletSymplecticHypersolver}
\begin{barticle}
\bauthor{\bsnm{Mathiesen}, \binits{F.B.}},
\bauthor{\bsnm{Yang}, \binits{B.}},
\bauthor{\bsnm{Hu}, \binits{J.}}:
\batitle{Hyperverlet: {{A Symplectic Hypersolver}} for {{Hamiltonian
  Systems}}}.
\bjtitle{Proceedings of the AAAI Conference on Artificial Intelligence}
\bvolume{36}(\bissue{4}),
\bfpage{4575}--\blpage{4582}
(\byear{2022})
\doiurl{10.1609/aaai.v36i4.20381}
\end{barticle}
\endbibitem

\bibitem[\protect\citeauthoryear{{Ober-Bl{\"o}baum} and
  Offen}{2023}]{Ober-Blobaum.Offen.2023.VariationalLearningEuler}
\begin{barticle}
\bauthor{\bsnm{{Ober-Bl{\"o}baum}}, \binits{S.}},
\bauthor{\bsnm{Offen}, \binits{C.}}:
\batitle{Variational learning of {{Euler}}--{{Lagrange}} dynamics from data}.
\bjtitle{J. Comput. Appl. Math.}
\bvolume{421},
\bfpage{114780}
(\byear{2023})
\doiurl{10.1016/j.cam.2022.114780}
\end{barticle}
\endbibitem

\bibitem[\protect\citeauthoryear{Offen and
  {Ober-Bl{\"o}baum}}{2024}]{Offen.Ober-Blobaum.2024.LearningDiscreteModels}
\begin{barticle}
\bauthor{\bsnm{Offen}, \binits{C.}},
\bauthor{\bsnm{{Ober-Bl{\"o}baum}}, \binits{S.}}:
\batitle{Learning of discrete models of variational {{PDEs}} from data}.
\bjtitle{Chaos}
\bvolume{34}(\bissue{1}),
\bfpage{013104}
(\byear{2024})
\doiurl{10.1063/5.0172287}
\end{barticle}
\endbibitem

\bibitem[\protect\citeauthoryear{Poli
  et~al.}{2020}]{Poli.Massaroli.ea.2020.HypersolversFastContinuousDepth}
\begin{bchapter}
\bauthor{\bsnm{Poli}, \binits{M.}},
\bauthor{\bsnm{Massaroli}, \binits{S.}},
\bauthor{\bsnm{Yamashita}, \binits{A.}},
\bauthor{\bsnm{Asama}, \binits{H.}},
\bauthor{\bsnm{Park}, \binits{J.}}:
\bctitle{Hypersolvers: {{Toward Fast Continuous-Depth Models}}}.
In: \bbtitle{Adv. Neural Inf. Process. Syst.},
vol. \bseriesno{33},
pp. \bfpage{21105}--\blpage{21117}.
\bpublisher{Curran Associates, Inc.}, \blocation{???}
(\byear{2020})
\end{bchapter}
\endbibitem

\bibitem[\protect\citeauthoryear{Qin
  et~al.}{2009}]{Qin.Guan.ea.2009.VariationalSymplecticAlgorithm}
\begin{barticle}
\bauthor{\bsnm{Qin}, \binits{H.}},
\bauthor{\bsnm{Guan}, \binits{X.}},
\bauthor{\bsnm{Tang}, \binits{W.M.}}:
\batitle{Variational symplectic algorithm for guiding center dynamics and its
  application in tokamak geometry}.
\bjtitle{Phys. Plasma}
\bvolume{16}(\bissue{4}),
\bfpage{042510}
(\byear{2009})
\doiurl{10.1063/1.3099055}
\end{barticle}
\endbibitem

\bibitem[\protect\citeauthoryear{Regazzoni et~al.}{2023}]{regazzoni2023latent}
\begin{botherref}
\oauthor{\bsnm{Regazzoni}, \binits{F.}},
\oauthor{\bsnm{Pagani}, \binits{S.}},
\oauthor{\bsnm{Salvador}, \binits{M.}},
\oauthor{\bsnm{Dede}, \binits{L.}},
\oauthor{\bsnm{Quarteroni}, \binits{A.}}:
Latent dynamics networks (ldnets): learning the intrinsic dynamics of
  spatio-temporal processes.
arXiv e-prints,
2305
(2023)
\end{botherref}
\endbibitem

\bibitem[\protect\citeauthoryear{Shen
  et~al.}{2020}]{Shen.Cheng.ea.2020.DeepEulerMethod}
\begin{botherref}
\oauthor{\bsnm{Shen}, \binits{X.}},
\oauthor{\bsnm{Cheng}, \binits{X.}},
\oauthor{\bsnm{Liang}, \binits{K.}}:
Deep {{Euler}} Method: Solving {{ODEs}} by Approximating the Local Truncation
  Error of the {{Euler}} Method.
arXiv
(2020)
\end{botherref}
\endbibitem

\bibitem[\protect\citeauthoryear{Tapley}{2024}]{tapley2024symplectic}
\begin{botherref}
\oauthor{\bsnm{Tapley}, \binits{B.K.}}:
Symplectic neural networks based on dynamical systems.
arXiv preprint arXiv:2408.09821
(2024)
\end{botherref}
\endbibitem

\bibitem[\protect\citeauthoryear{Tenachi et~al.}{2023}]{tenachi2023deep}
\begin{barticle}
\bauthor{\bsnm{Tenachi}, \binits{W.}},
\bauthor{\bsnm{Ibata}, \binits{R.}},
\bauthor{\bsnm{Diakogiannis}, \binits{F.I.}}:
\batitle{Deep symbolic regression for physics guided by units constraints:
  toward the automated discovery of physical laws}.
\bjtitle{The Astrophysical Journal}
\bvolume{959}(\bissue{2}),
\bfpage{99}
(\byear{2023})
\end{barticle}
\endbibitem

\bibitem[\protect\citeauthoryear{Zhu and Li}{2024}]{zhu2024dyngma}
\begin{barticle}
\bauthor{\bsnm{Zhu}, \binits{A.}},
\bauthor{\bsnm{Li}, \binits{Q.}}:
\batitle{{DynGMA: A robust approach for learning stochastic differential
  equations from data}}.
\bjtitle{J. Comput. Phys.}
\bvolume{513},
\bfpage{113200}
(\byear{2024})
\end{barticle}
\endbibitem

\end{thebibliography}

\end{document}